%% file: main.tex
\documentclass{article} %
\ifdefined\pdfsuppressptexinfo\pdfsuppressptexinfo=-1\fi %
\usepackage{iclr2027_conference,times}
\usepackage[T1]{fontenc}

\input{math_commands.tex}

\usepackage{hyperref}
\usepackage{url}
\usepackage{booktabs}
\usepackage{multirow}
\usepackage{subcaption}
\usepackage{graphicx}
\usepackage{wrapfig}
\usepackage{needspace}
\usepackage{xcolor}
\usepackage{colortbl}
\usepackage{amsmath,amssymb}
\usepackage{enumitem}
\usepackage{listings}
\usepackage[most]{tcolorbox}

\lstdefinestyle{prompt}{
    basicstyle=\ttfamily\fontsize{7pt}{8pt}\selectfont,
    frame=none,
    breaklines=true,
    backgroundcolor=\color{lightgray},
    breakatwhitespace=true,
    breakindent=0pt,
    escapeinside={(*@}{@*)},
    numbers=none,
    numbersep=5pt,
    xleftmargin=5pt,
}
\tcbset{
  aibox/.style={
    top=10pt,
    colback=white,
    colframe=black,
    colbacktitle=black,
    enhanced,
    center,
    attach boxed title to top left={yshift=-0.1in,xshift=0.15in},
    boxed title style={boxrule=0pt,colframe=white,},
  }
}
\newtcolorbox{AIbox}[2][]{aibox, title=#2,#1}

\newcommand{\todo}[1]{\textcolor{red}{[TODO: #1]}}

\newcommand{\note}[1]{\textcolor{blue}{[#1]}}

\newcommand{\hide}[1]{}
\newif\ifshowrevision
\showrevisiontrue
\newif\ifpreprint
\preprinttrue %
\ifpreprint\showrevisionfalse\fi
\newcommand{\rev}[1]{\ifshowrevision{\color{blue}#1}\else#1\fi}
\newcommand{\revall}{\ifshowrevision\color{blue}\fi}
\newcommand{\bench}{\textsc{KernelShift}}

\title{\bench{}: Benchmarking Kernel \\Optimization Across Evolving Hardware}

\ifpreprint
\author{Shuaiwei Huang$^{1*}$ \quad Weihua Du$^{1*}$ \quad Allen Wu$^{1*}$ \quad Ivan Fox$^{2}$ \\ \textbf{Manupa Karunaratne$^{2}$ \quad Yiming Yang$^{1}$ \quad Sean Welleck$^{1}$} \\[4pt]
{\normalfont $^{1}$Carnegie Mellon University \quad $^{2}$Arm Ltd.} \\
{\normalfont\small\texttt{\{shuaiweh,weihuad,allenwu,yiming,swelleck\}@andrew.cmu.edu}} \\
{\normalfont\small\texttt{\{Ivan.Fox,Manupa.Karunaratne\}@arm.com}}}
\iclrfinalcopy
\else
\author{Anonymous authors \\ Paper under double-blind review}
\fi

\begin{document}

\maketitle
\ifpreprint\lhead{Preprint}
{\renewcommand{\thefootnote}{}\footnotetext{$^{*}$Equal contribution. Project website: \url{https://cmu-l3.github.io/kernel-shift/}}}
\fi

\input{sections/00_abstract2}
\input{sections/01_introduction2}
\input{sections/03_benchmark2}

\input{sections/04_setup}

\input{sections/05_results}
\input{sections/02_related_work}
\input{sections/07_conclusion}

\input{sections/99_statements}

\bibliography{references}
\bibliographystyle{iclr2027_conference}

\appendix
\input{sections/appendix}

\end{document}

%% file: math_commands.tex
\usepackage{amsmath,amsfonts,bm}

\def\eqref#1{equation~\ref{#1}}

\def\1{\bm{1}}

\DeclareMathAlphabet{\mathsfit}{\encodingdefault}{\sfdefault}{m}{sl}
\SetMathAlphabet{\mathsfit}{bold}{\encodingdefault}{\sfdefault}{bx}{n}

%% file: sections/00_abstract2.tex
\begin{abstract}
Efficient AI computation depends on hardware that \textit{evolves over time}, shifting the strategies needed to implement performant code. 
For CPU kernels, a key evolving factor is the set of available \textit{architectural extensions}, which expose specialized hardware instructions that change the techniques used to optimize kernels.
We study how effectively coding agents use these new, potentially unfamiliar extensions to write fast code.
\rev{We introduce \textbf{\bench{}}, a benchmarking pipeline and a set of environments for evaluating coding agents in kernel optimization across evolving CPU architectures.}
Its agent-assisted, human-verified pipeline turns code repositories into benchmark instances, and its environments let agents optimize code with feedback from the target hardware.
Using this pipeline,
we construct \rev{99 kernels} from real-world, actively developed repositories (ncnn, llama.cpp, Arm SIMD Loops, and KleidiAI), including inference workloads derived from real model executions.
\bench{} supports incorporating new kernels and hardware configurations as libraries and processors evolve.
We evaluate agents with proprietary and open-weight models across progressively newer architectural extensions (Arm Neon, SVE, SVE2, and SME2), examining how performance changes as hardware evolves.
\rev{On established extensions, agents beat the auto-vectorized compiler everywhere ($2.4$--$19.8\times$) and exceed general-purpose libraries such as ncnn (getting up to $2.74\times$ over its Neon kernels), but fall behind Arm's hand-tuned KleidiAI kernels ($0.70\times$).} In end-to-end inference of two LLMs, agent-written kernels achieve a $1.3$--$1.6\times$ prefill speedup over llama.cpp's fast path. 
However, current agents do not yet reliably adapt to new extensions that require unfamiliar programming models, such as SME2's matrix tiles.
\end{abstract}

%% file: sections/01_introduction2.tex
\section{Introduction}
Modern ML workloads run across an increasingly diverse range of hardware, from CPUs to GPUs with different architectures and capabilities~\citep{reddi2020mlperf}. %
As these platforms continue to evolve, the optimization strategies needed to use them efficiently evolve as well.
Meanwhile, AI coding agents are increasingly being used to optimize performance-critical code.
For example, frontier labs such as OpenAI report
using AI models to autonomously rewrite and optimize production GPU kernels~\citep{openai2026gpt56,novikov2025alphaevolve}.
As AI agents take on more performance-critical optimization, an important question is whether they can adapt their optimization strategies as the underlying hardware \textit{evolves over time}.

In CPUs, hardware evolution often takes the form of new \textit{architectural extensions}, which introduce new instructions and expose them to programmers through a new interface.
Extensions can improve performance, but exploiting them often requires fundamentally restructuring a kernel's implementation.
For example, Arm's matrix extensions support accumulating products directly into 2D matrix tiles, enabling implementations of matrix multiplication that differ from those based on earlier 1D vector instructions~\citep{arm2026acle,remke2024hellosme}.
An agent can therefore produce correct but suboptimal code if it relies on older vector instructions when newer matrix instructions would be advantageous.

\begin{figure}[t]
  \centering
  \includegraphics[width=\linewidth]{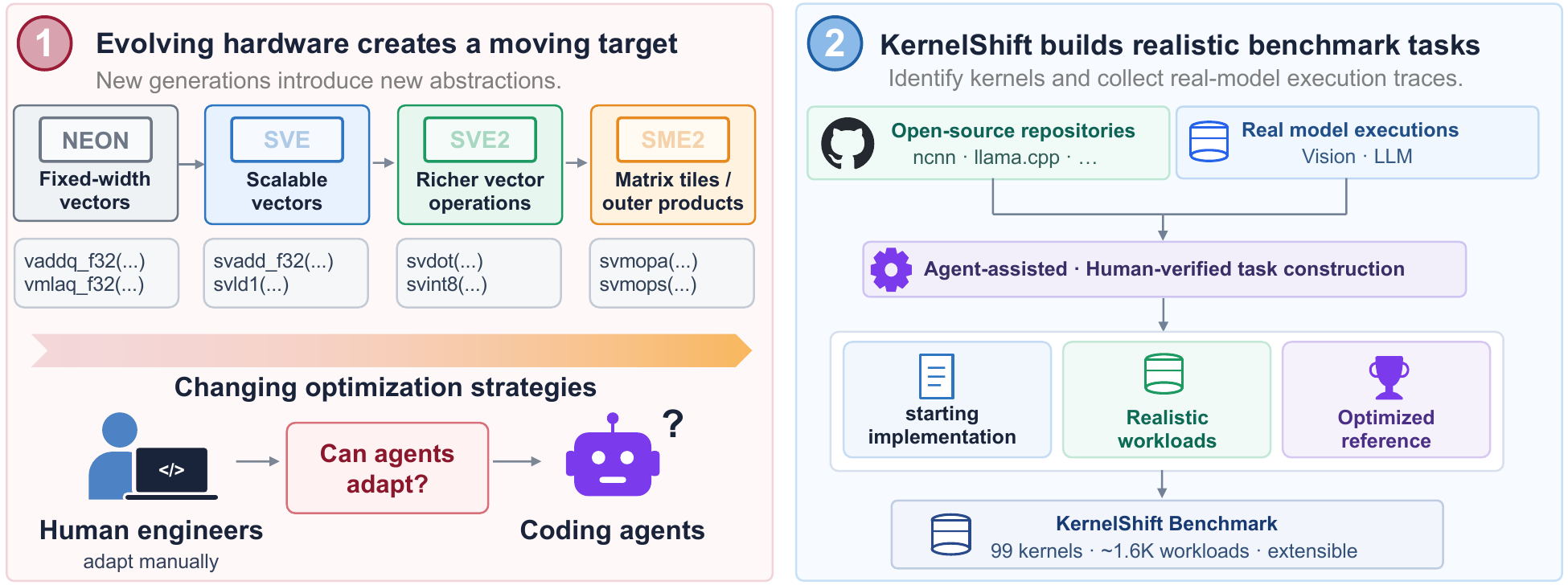}
  \caption{\textbf{Overview of \bench{}.}
\textbf{(Left)} Evolving hardware changes the space of performant kernel implementations,
requiring both human engineers and coding agents to continually adapt their strategies.
\textbf{(Right)} To study this adaptation, \bench{} constructs benchmark tasks from real-world sources and model executions using an agent-assisted, human-verified pipeline.}
\vspace{-0.6cm}
  \label{fig:concept}
\end{figure}

To study kernel optimization under evolving hardware, we introduce \bench{} (\autoref{fig:concept} (left)), a benchmarking pipeline and a set of environments for evaluating AI coding agents across evolving CPU architectures.
The key idea is to evaluate how agents optimize matched kernels, where coverage permits, across successive \textit{architectural extensions} that shift the space of performant implementations.
Specifically, we evaluate agents across Arm's vector and matrix extensions (i.e., Neon, SVE, SVE2, and SME2), which provide instructions for parallelizing computations common in machine learning workloads~\citep{stephens2017sve,arm2026acle}.
Our work builds on a growing literature on hardware-aware kernel optimization~\citep{he2025simdbench,wen2025multikernelbench,zang2026kernelgenbench}, with a specific focus on sequences of hardware changes that expose progressively newer programming interfaces.

We design an agent-assisted, human-verified pipeline with source-specific adapters for transforming code repositories into benchmark instances (\autoref{fig:concept} (right)), together with environments that let agents iteratively optimize code using compilation and performance feedback from the target hardware (\autoref{fig:architecture}). Using this pipeline, we construct \rev{99 kernel specifications} from real-world, actively developed repositories, including workloads derived from real model executions. The pipeline and environments are designed to incorporate new kernels and hardware configurations as libraries and processors evolve.

We evaluate coding agents with proprietary and open-weight models across progressively newer Arm architectural extensions. 
We measure both optimization performance and their use of newly available instructions, comparing their implementations against compiler-optimized code, production-library implementations, and expert kernels where available. \rev{Agents readily find headroom on established vector extensions: on SVE, they beat the auto-vectorized compiler in every source family and exceed general-purpose libraries such as ncnn (getting up to \(2.74\times\) over its Neon kernels within 40 tool calls), but fall behind Arm's hand-tuned KleidiAI kernels (\(0.70\times\)).} However, this does not reliably carry over to newer programming models: on SME2, agents end up far slower than the expert kernels.

Our main contributions are:
 \begin{itemize}[leftmargin=*, itemsep=1pt, topsep=2pt, parsep=2pt]
    \item \textbf{\bench{}}, a benchmarking pipeline and a set of environments to evaluate coding agents on evolving hardware and programming interfaces. The agent-assisted, human-verified pipeline supports adding kernels and hardware configurations as software libraries and processors evolve.
    \item \textbf{A benchmark of 99 real-world CPU kernels} built with this pipeline, including workloads derived from real model executions.
    \item \textbf{An empirical study of coding agents across Arm architecture extensions}, measuring both optimization performance and agents' ability to exploit newly available hardware instructions.
\end{itemize}

%% file: sections/03_benchmark2.tex
\section{\bench{}}
\label{sec:benchmark}

Our goal is to evaluate agents' ability to write performant code for \textit{evolving hardware}. 
We introduce our problem setting, motivate our use of \textit{architectural extensions} to distinguish stages of CPU hardware evolution, and describe our automated pipeline for collecting benchmark instances.

\paragraph{Hardware-aware kernel optimization.}
We consider the setting of \textit{hardware-aware kernel optimization}.
In this setting, a \textit{kernel} is a program that performs a specific computation, such as a matrix multiplication.
Each task consists of an initial kernel implementation $x$, together with context $c$.
The context includes the function signature, symbolic tensor shapes, and data types.
Concrete workload dimensions and input values remain hidden
(Appendix~\ref{app:interface}).

The kernel is executed on target hardware, $h=(I,m)$.
Here, $I$ denotes the target \textit{instruction-set architecture} (ISA), including base instructions and those introduced by supported \textit{architectural extensions}.
Programmers can access specialized instructions through \textit{intrinsics}: compiler-supported operations exposed in C/C++.
The \textit{microarchitecture} $m$ is a concrete processor design that implements $I$ on physical hardware.
In general, the goal is to develop an agent $A$ that takes in an initial implementation, context, and hardware, and produces an optimized implementation $x'=A(x,c,h)$:
\begin{equation}
\label{eqn:objective}
\max_A \mathbb{E}_{(x,c)\sim\mathcal{D},\,h\sim\mathcal{H}}
\left[s\bigl(x,A(x,c,h)\bigr)\right],
\end{equation}
where $\mathcal{D}$ is a distribution of optimization tasks and $\mathcal{H}$ is a distribution of hardware configurations.

For a fixed task and target hardware, the score $s(x,x')$ is the
geometric mean speedup over the task's optimized reference across $n$
hidden workloads, which specify input configurations such as matrix dimensions.
That is, let $t_i^{\mathrm{ref}}$ and $t_i'$ denote the runtimes of the
optimized reference and the agent implementation on workload $i$.
The score is:
\begin{equation}
s = \left(\prod_{i=1}^{n}\frac{t_i^{\mathrm{ref}}}{t_i'}\right)^{1/n},
\end{equation}
with $s=0$ if the agent implementation fails any correctness
check (Appendix~\ref{app:correctness}).

\paragraph{Evolving hardware.}
Rather than sampling hardware configurations to approximate \autoref{eqn:objective}, we are interested in sequences of hardware configurations that introduce new functionality to a common base architecture,
\begin{equation}
h_1 \longrightarrow h_2 \longrightarrow \cdots \longrightarrow h_T.
\end{equation}
The kernel's desired functionality remains fixed across the sequence, while the implementations that realize the functionality efficiently may change substantially.

Specifically, we primarily focus on changes to the available instruction set $I$ through new extensions, since new extensions can introduce different ways to optimize computations and hence cause large shifts in the space of performant programs.
For example, as illustrated in \autoref{fig:concept}, Arm's \textit{Neon} extension exposes fixed-width vector operations, while the later \textit{scalable vector extensions} (SVE, SVE2) support code that adapts to different vector widths and provide additional vector operations.
The more recent \textit{scalable matrix extension} (SME) introduces dedicated matrix storage and instructions for accumulating vector outer products into matrix tiles, enabling new ways to implement matrix multiplication.

Motivated by these architectural shifts, we evaluate agents' kernel optimization performance across such evolving extensions. In particular, we investigate whether the choice of extension impacts agent performance, and whether agents can effectively leverage newer intrinsics that are underrepresented in their training data. Therefore, we construct an automated benchmark framework that seamlessly accommodates new extensions as they arise. The benchmark design and automated generation techniques are detailed next.

\subsection{Agent-Assisted Task Construction}
\label{sec:pipeline}

Our construction framework turns performance-critical kernels from open-source projects into
benchmark tasks. For each kernel, it constructs a portable C++ starting implementation $x$, the
context $c$ exposed to the agent, hidden workloads, and, when available,
an optimized reference. Tasks are constructed by coding agents following fixed procedures
backed by deterministic generation and validation scripts. Human input is limited to a few
decisions: selecting kernels, approving workloads where no model execution can be traced, and
reviewing the resulting expert references.
Appendix~\ref{sec:app-construction} documents the complete construction process.

\paragraph{Kernel and reference construction.}
For each source repository, a construction agent follows a fixed procedure~(Appendix~\ref{app:skills}) to identify candidate
kernel computations and extract their fixed configuration parameters and variable workload
dimensions. It then constructs a portable C++ starting implementation using simple loop-based
code. When the repository contains an optimized implementation of the same kernel, the agent
also extracts it as the optimized reference.

\paragraph{Workload construction.}
To capture realistic workloads, we derive input configurations from real model executions
whenever possible. For model-derived kernels, the construction agent first identifies the source
model in which the kernel appears. Fixed configuration parameters then come from this model,
while dimensions are collected from execution traces. For example, ncnn
workloads come from vision-model executions, while llama.cpp workloads are collected during
LLM inference on ShareGPT prompts~\citep{sharegpt}. We keep the observed tensor shapes rather than the original
activation values, under the assumption that kernel performance is driven primarily by tensor
dimensions and operator configuration rather than the specific numerical values of the activations.
We then deduplicate and subsample the collected shapes to form the hidden evaluation workloads.
For kernels without a clearly identifiable source model, we instead synthesize workloads in two
groups: correctness-oriented cases that cover corner cases, and performance-oriented cases that
probe efficiency over representative problem sizes. More details appear in
Appendix~\ref{app:workloads}.

\paragraph{Finalizing benchmark tasks.}
Finally, we package the starting implementation, agent-visible context, evaluation
workloads, and the optimized reference from the source library, when available, into a
self-contained benchmark task. Each task is validated by compiling and running the starting
implementation on the target hardware before it is included in \bench{}.

\subsection{Benchmark Composition and Coverage}
\label{sec:kernels}
We apply the construction framework from
Section~\ref{sec:pipeline} to several widely used, performance-critical
open-source repositories.

\paragraph{Composition.}
\bench{} currently contains kernel tasks derived from four sources
(Table~\ref{tab:composition}): ncnn~\citep{ncnn}, llama.cpp~\citep{gerganov2023llamacpp},
Arm SIMD Loops~\citep{armsimdloops}, and KleidiAI~\citep{kleidiai}.
Together, they cover computer-vision and LLM inference kernels,
general SIMD workloads, and vendor-optimized microkernels across a range of
data types and optimization patterns.
The ncnn and llama.cpp tasks are derived from 11 source models, providing
kernel configurations representative of real model architectures.
Appendices~\ref{app:composition} and~\ref{app:coverage} provide the full
composition and per-generation coverage.

\begin{table}[t]
\caption{\textbf{Composition of \bench{}.} We apply our kernel collection pipeline to common or vendor-provided repositories, resulting in a wide range of kernels.}
\label{tab:composition}
\centering
\small
\resizebox{\linewidth}{!}{%
\begin{tabular}{llllr}
\toprule
Source & Op types & Precisions & Expert reference & \#Kernels \\
\midrule
ncnn       & conv, dwconv, pool & FP32, INT8 (w8a8ch)        & ncnn Neon & 23 \\
llama.cpp  & gemm, moe, attn, rmsnorm & FP32, BF16, q8\_0, q4\_K\_M & ggml      & 26 \\
SIMD Loops & loops (reduction, string, search, \ldots) & fp/int/uint     & Arm SVE   & 24 \\
KleidiAI   & gemm, conv, dwconv (NHWC) & FP32, FP16, BF16, INT8, INT4, INT2 & KleidiAI  & \rev{26 (6 SVE, 20 SME2)} \\
\midrule
Total & & & & \rev{99} \\
\bottomrule
\end{tabular}}
\end{table}

\paragraph{Workloads.}
\bench{} contains approximately 1.6K workloads across the 99 kernel specifications, constructed as
described in Section~\ref{sec:pipeline} and spanning a broad range of tensor dimensions and
sequence lengths.

\paragraph{Hardware and reference coverage.}
The 99 kernels form the union across targets: every task can be instantiated on each target, but
an optimized reference is not available for every source on every generation, so each generation
uses the tasks that have one. For each task, the reference is a strong production or expert
implementation available on the target (Appendix~\ref{app:coverage}).

\subsection{Evaluation Environment}
\label{sec:environment}
\begin{figure}[t]
  \centering
  \vspace{-2mm}
  \includegraphics[width=\linewidth]{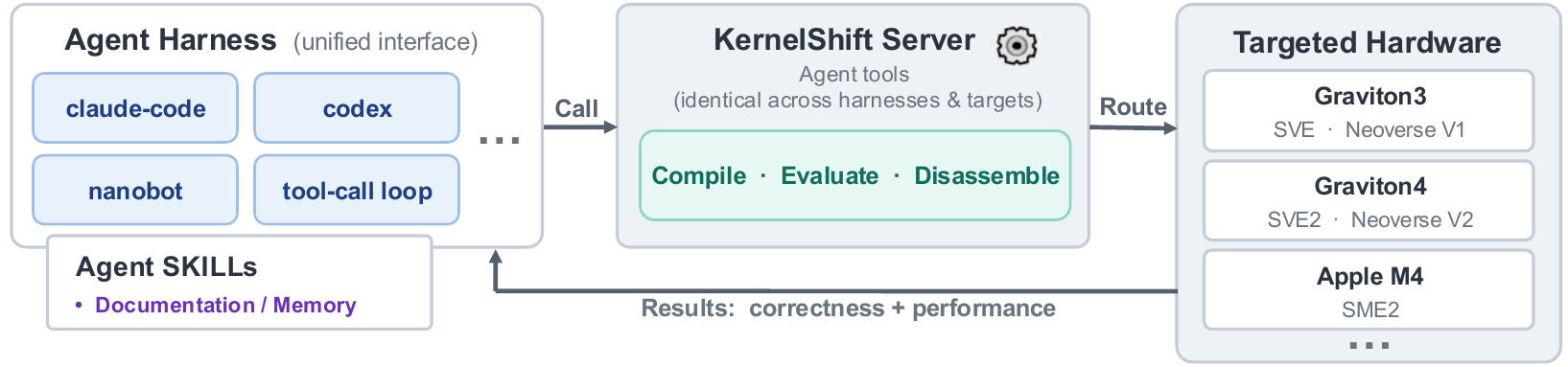}
\caption{\textbf{Evaluation architecture of \bench{}.}
A unified server exposes the same tools and skills to all agent harnesses and routes their requests to evaluators running on the target hardware.}
  \label{fig:architecture}
\end{figure}

\bench{} uses a centralized control plane to run agent harnesses and coordinate evaluation
across distributed target machines (\autoref{fig:architecture}).
\paragraph{Hardware-in-the-loop evaluation.}
Each target platform is backed by an evaluator running directly on its hardware. When an agent
calls a tool, the request is routed to this evaluator, which compiles and executes the candidate on
the target machine and returns the feedback, so that correctness and performance measurements
reflect the hardware being evaluated. This design decouples the agent harness from
target-specific execution, allowing the same agent setup to be used across hardware platforms.
Appendix~\ref{app:infrastructure} gives the deployment configurations.
\hide{
Each hardware generation $g$ is backed by a dedicated evaluator running on its
designated target machine $m_g$: SVE on Graviton3, SVE2 on Graviton4,
and SME2 on Apple M4.
As illustrated in Figure~\ref{fig:architecture}, a central control plane provisions
the target environment, dispatches $(K,g)$ tasks, and collects the resulting agent
trajectories.
Candidate implementations are compiled and executed directly on the target machine,
so both correctness and performance feedback reflect the hardware generation being evaluated.
The evaluator verifies the target ISA before execution and applies the corresponding
compilation target for each generation.
}

\paragraph{Unified agent interface.}
\label{sec:interface}
To compare different agent systems under the same environment,
\bench{} exposes a harness-agnostic interface through which all agents access the benchmark.
The interface provides the same three benchmark tools to every harness:
\emph{compile}, \emph{evaluate}, and \emph{disassemble}.
These tools allow agents to iteratively refine their solutions with
correctness, performance, and low-level instruction feedback from the target hardware.
Appendix~\ref{app:prompts} provides the complete prompts, skills, and tool schemas.

\paragraph{Measurement and integrity.}
\label{sec:protocol}
Correctness is checked before performance is measured, and both agent implementations
and comparison baselines are tested on the same target machine under the same timing
protocol. We use wall-clock execution time as the common performance measure across
platforms, since hardware cycle counters are not consistently available on all target systems.
Performance is reported as the speedup of a kernel over its expert reference on the same machine.
Measurements use repeated runs on a
single core, and each evaluation runs in an isolated process to prevent interference between
evaluations. We also reject implementations that violate the benchmark constraints, such as
using multithreading. Full timing, isolation, and integrity details are provided in
Appendix~\ref{app:measurement}.

\input{tables/main-results}

%% file: tables/main-results.tex
\begin{table}[t]
\caption{\textbf{Results on SVE (Graviton3)}. Each cell reports the geometric mean over kernels of
Speedup@40 against the source's expert reference and, in parentheses, the fraction of kernels that
beat references ($>\!1$). The best speedup in each column is in bold. \rev{Each row is an end-to-end
system (harness and model) run under the same limits, so rows compare systems rather than models alone.}}
\label{tab:main-sve}
\centering\small
\resizebox{\linewidth}{!}{%
\begin{tabular}{llcccc}
\toprule
Harness & Model & \multicolumn{4}{c}{Geomean Speedup@40 \ (beat-ref \%) vs.\ expert reference} \\
\cmidrule(lr){3-6}
 & & ncnn (23) & llama.cpp (26) & SIMD Loops (24) & KleidiAI (6) \\
\midrule
Claude Code & Claude Opus 4.8   & 2.574\,{\footnotesize(96\%)} & 1.779\,{\footnotesize(73\%)} & \textbf{1.555}\,{\footnotesize(96\%)} & \textbf{0.701}\,{\footnotesize(33\%)} \\
Claude Code & Claude Sonnet 4.6 & 1.934\,{\footnotesize(83\%)} & 1.243\,{\footnotesize(46\%)} & 1.516\,{\footnotesize(100\%)} & 0.386\,{\footnotesize(33\%)} \\
Codex       & GPT-5.6-sol       & 1.815\,{\footnotesize(78\%)} & 1.297\,{\footnotesize(62\%)} & 1.487\,{\footnotesize(96\%)} & 0.431\,{\footnotesize(17\%)} \\
Codex       & GPT-5.6-luna      & 0.674\,{\footnotesize(35\%)} & 0.627\,{\footnotesize(27\%)} & 1.127\,{\footnotesize(79\%)} & 0.147\,{\footnotesize(0\%)} \\
nanobot     & Claude Opus 4.8   & 2.597\,{\footnotesize(91\%)} & 1.708\,{\footnotesize(73\%)} & 1.521\,{\footnotesize(96\%)} & 0.453\,{\footnotesize(17\%)} \\
nanobot     & Claude Sonnet 4.6 & 2.046\,{\footnotesize(87\%)} & 1.449\,{\footnotesize(62\%)} & 1.503\,{\footnotesize(96\%)} & 0.231\,{\footnotesize(0\%)} \\
nanobot     & GPT-5.6-sol       & 2.156\,{\footnotesize(87\%)} & 1.334\,{\footnotesize(58\%)} & 1.543\,{\footnotesize(100\%)} & 0.319\,{\footnotesize(0\%)} \\
nanobot     & GPT-5.6-luna      & 1.110\,{\footnotesize(57\%)} & 0.878\,{\footnotesize(31\%)} & 1.182\,{\footnotesize(79\%)} & 0.161\,{\footnotesize(0\%)} \\
nanobot     & DeepSeek-V4-Pro   & 1.091\,{\footnotesize(57\%)} & 0.850\,{\footnotesize(46\%)} & 1.458\,{\footnotesize(88\%)} & 0.231\,{\footnotesize(17\%)} \\
nanobot     & GLM-5.3           & \textbf{2.743}\,{\footnotesize(91\%)} & \textbf{1.954}\,{\footnotesize(85\%)} & 1.545\,{\footnotesize(96\%)} & 0.499\,{\footnotesize(17\%)} \\
tool-call loop & Claude Opus 4.8   & 2.308\,{\footnotesize(83\%)} & 1.828\,{\footnotesize(73\%)} & 1.480\,{\footnotesize(92\%)} & 0.496\,{\footnotesize(33\%)} \\
tool-call loop & Claude Sonnet 4.6 & 2.043\,{\footnotesize(87\%)} & 1.915\,{\footnotesize(81\%)} & 1.521\,{\footnotesize(100\%)} & 0.297\,{\footnotesize(17\%)} \\
tool-call loop & GPT-5.6-sol       & 1.999\,{\footnotesize(91\%)} & 1.671\,{\footnotesize(81\%)} & 1.513\,{\footnotesize(96\%)} & 0.307\,{\footnotesize(33\%)} \\
tool-call loop & GPT-5.6-luna      & 0.955\,{\footnotesize(48\%)} & 0.854\,{\footnotesize(31\%)} & 1.245\,{\footnotesize(75\%)} & 0.164\,{\footnotesize(17\%)} \\
\midrule
\multicolumn{2}{l}{single-shot, no tools (Claude Opus 4.8)$^{\dagger}$} & 0.577\,{\footnotesize(35\%)} & 0.324\,{\footnotesize(8\%)} & 0.852\,{\footnotesize(42\%)} & 0.152\,{\footnotesize(0\%)} \\
\multicolumn{6}{l}{\footnotesize$^{\dagger}$For correctness failures, we substitute the starter-kernel score.} \\
\midrule
\multicolumn{6}{l}{\emph{Reference tiers}} \\
\multicolumn{2}{l}{scalar + \texttt{-O2} (no vectorization)} & 0.190\,{\footnotesize(0\%)} & 0.090\,{\footnotesize(8\%)} & 0.248\,{\footnotesize(0\%)} & 0.025\,{\footnotesize(0\%)} \\
\multicolumn{2}{l}{scalar + \texttt{-O2} (Neon auto-vectorized)} & 0.191\,{\footnotesize(0\%)} & 0.117\,{\footnotesize(12\%)} & 0.398\,{\footnotesize(17\%)} & 0.031\,{\footnotesize(0\%)} \\
\multicolumn{2}{l}{scalar + \texttt{-O3}, SVE enabled} & 0.229\,{\footnotesize(9\%)} & 0.126\,{\footnotesize(15\%)} & 0.468\,{\footnotesize(25\%)} & 0.035\,{\footnotesize(0\%)} \\
\bottomrule
\end{tabular}}
\end{table}

%% file: sections/04_setup.tex
\section{Experimental Setup}
\label{sec:setup}

\paragraph{Models and harnesses.}
We evaluate six frontier language models: Claude Opus 4.8~\citep{anthropic2026opus48} and Claude Sonnet 4.6~\citep{anthropic2026sonnet46}, GPT-5.6-sol and GPT-5.6-luna~\citep{openai2026gpt56}, DeepSeek-V4-Pro~\citep{deepseek2026v4}, and GLM-5.3~\citep{zhipu2026glm}. We pair these models with four harnesses: Claude Code~\citep{anthropic2025claudecode}, Codex~\citep{openai2025codexcli}, nanobot~\citep{nanobot}, 
and a minimal tool-call loop. We evaluate Claude Code and Codex with their own models, while nanobot provides a common harness across all six models to disentangle model and harness effects. The end-to-end study (Section~\ref{sec:e2e}) additionally uses Claude Fable 5.1~\citep{anthropic2026fable51} with Claude Code. Appendix~\ref{app:harness} contains full model and harness configurations. 

\paragraph{Task sets.}
Our experiments draw from the 99-kernel benchmark, with generation-specific coverage:
The SVE comparison contains 79 kernels (23 ncnn, 26 llama.cpp, 24 SIMD Loops,
and 6 KleidiAI). We use a 32-kernel subset for ablations and controlled studies. 

\paragraph{Hardware and budgets.}
We evaluate SVE on Graviton3 (AWS c7g), SVE2 on Graviton4 (AWS c8g), and SME2
on Apple M4. Each run receives a budget of $T=40$ tool calls, with \emph{compile},
\emph{evaluate}, and \emph{disassemble} calls all counting toward the budget.
For test-time scaling experiments, we extend the budget to 100 tool calls.

\paragraph{Metrics.}
Our primary metric, Speedup@40, is the score $s$ (Section~\ref{sec:benchmark}) of the best
correct implementation within 40 calls, averaged geometrically over kernels. We also report the
fraction of kernels that beat the reference and compiler baselines, and performance versus budget.

%% file: sections/05_results.tex
\section{Results}
\label{sec:results}

\subsection{Agents find substantial speedups}
\label{sec:competitive}
\label{sec:rq1}

Strong coding agents can exploit substantial optimization headroom in production kernels. On SVE, the best systems reach 2.74$\times$ the ncnn reference,
1.95$\times$ the llama.cpp reference, and 1.56$\times$ the SIMD Loops reference
(Table~\ref{tab:main-sve}). In contrast, compiler-based automatic optimization remains far
behind expert references, with even \texttt{-O3} and SVE reaching at most 0.47$\times$. The
\texttt{-O3}, SVE-enabled build is also the code that each agent starts from. Every system is faster than this \texttt{-O3} reference across all four repositories ($2.4$--$19.8\times$; Appendix~\ref{app:baselines}).

Highly tuned vendor kernels remain substantially more competitive. On the six KleidiAI
kernels with SVE expert references, no evaluated system matches the vendor implementations
in geometric mean; the best, Claude Code with Claude Opus 4.8, reaches 0.70$\times$, beating
the reference on two kernels but falling behind on the other four. Agents exploit headroom in general-purpose production
    libraries, but matching specialized vendor kernels is harder.

\input{tables/ladder-evolve}
\subsection{Adapting to Evolving Hardware}
\label{sec:adapt}
\label{sec:rq3}

\paragraph{Agents struggle more with the programming-model shift to SME2.}
\rev{The biggest drop appears on SME2, which introduces streaming mode and ZA
matrix tiles instead of just adding more vector instructions. On the same KleidiAI
kernels, performance relative to the expert reference falls from SVE to SME2 for
three of the five models: the SME2-to-SVE ratios are $0.42$, $0.46$, and $0.41$ for
DeepSeek-V4-Pro, GLM-5.3, and GPT-5.6-luna, while the performance of the Claude models remains
similar (Table~\ref{tab:evolve}). GLM-5.3, the strongest of the five models on
SVE, falls to less than half, so doing well on SVE does not carry over when a new extension
requires a different programming style. 

We also run the fixed-machine ladder on Apple M4, with GPT-5.6-luna as the base model and nanobot as harness. Over three seeds, exposing streaming SVE lowers Speedup@40 from $0.141$
(Neon only) to $0.059$, and exposing full SME2 recovers part of the loss, to $0.085$ (Table~\ref{tab:isa-ladder}). 
Neon > SME2 > streaming SVE holds in every seed. The performance gap mostly comes from the different optimization paths that the agent applied. With streaming SVE available, the agent almost abandons Neon and settles on streaming SVE or falls back to plain C if compilation or correctness errors repeatedly happen. %
With full SME2 available, the recovery over streaming SVE does not come from ZA outer products: the agent simply keeps a Neon kernel more often than with streaming SVE only (Appendix~\ref{app:ladder-simd-class}). None of the final SME2-tier kernels use ZA outer products, and across all SME2 runs the agent writes ZA code only twice, with neither version passing. Streaming and ZA code also fail to compile more often: in the first seed, 13\% of compiles fail on the Neon tier, 20\% with streaming SVE and 24\% with full SME2 (Appendix~\ref{app:ladder-compile}).}

\paragraph{SME2 Case Study.} On \texttt{gemm\_bf16\_n512\_k512}, Claude Opus 4.8 switches to streaming
mode and uses MOPA/ZA accumulation, reaching $0.94\times$ the expert
implementation. Claude Sonnet 4.6 and DeepSeek-V4-Pro, in contrast, stay with
older vector-style implementations and reach only $0.06\times$ and
$0.05\times$. The same holds across kernels: nearly all
final SVE solutions use SVE intrinsics, but of the 18 final SME2 solutions on the six KleidiAI kernels
of Table~\ref{tab:evolve} (three seeds), only 15 (Opus), 5 (Sonnet), 5 (GLM-5.3), and 1 (DeepSeek-V4-Pro)
use the ZA tiles (Appendix~\ref{app:intrinsic-usage}).
Appendix~\ref{app:sme2-case} traces in detail a representative SME2 optimization
trajectory.

\paragraph{Agents adapt more readily to incremental ISA extensions.}
By contrast, the move from SVE to SVE2 costs much less. On the eight SIMD Loops kernels whose
expert SVE2 implementation needs SVE2-only instructions, the SVE2-to-SVE ratios are between $0.94$
and $1.05$ for every model (Table~\ref{tab:evolve}), and most
final SVE2 solutions of the Claude models use the expert's SVE2-only instructions
(Appendix~\ref{app:intrinsic-usage}).
The fixed-machine instruction ladder on Graviton4 shows the same pattern:
Neon and SVE2 reach $1.105\times$ and $1.101\times$ (Table~\ref{tab:isa-ladder}).
The main gap comes from how agents implement kernels on Graviton4's 128-bit SVE vectors:
about half of the gap comes from runs where the agent never got SVE code past the correctness
check and fell back to scalar code. The other half comes from SVE code that is correct but
slower, %
where the agent did not use the right SVE instructions
(Appendix~\ref{app:isa-ladder}).
This is a milder form of the SME2 failures above: when agents do not use a newer programming
interface well, their kernels end up slower than plain Neon code.

\hide{\todo{Prose for Table~\ref{tab:evolve}: lead with SIMD Loops SVE$\to$SVE2 (the controlled
same-kernel, dual-expert comparison) as the primary adaptation result; present KleidiAI SVE$\to$SME2
as complementary, noting that agents stay below the vendor kernel on both ISAs, so its ratio mainly
rules out ``SME2 is intrinsically harder'' as the explanation for the earlier KleidiAI-vs-library gap
rather than establishing an adaptation gap. Contrast with Table~\ref{tab:isa-ladder}, which isolates
the instruction set on a \emph{fixed} machine against a compiler reference, whereas Table~\ref{tab:evolve}
uses \emph{cross-machine} runs against upgraded \emph{expert} baselines.}

\todo{Read Table~\ref{tab:isa-ladder} together with S2c (SVE restricted to a 128-bit vector length):
the ladder isolates the instruction set, S2c isolates the vector width, and L2 against the SVE cells
of Table~\ref{tab:main-sve} isolates the microarchitecture, which together decompose a
generation-to-generation difference into its three causes.}
\todo{Portable tier (S3): how much of the gain comes from SIMD and how much from loop and memory layout.
Add the disassembly-based measurement of how often agents actually emit generation-specific
instructions, and repeat the SME2 campaign with more models once the task set is fixed.}}

\subsection{Agent Gains Depend on Expert Headroom}
\label{sec:where}
\label{sec:rq2}
\begin{figure}[t]
\centering
\begin{minipage}[t]{0.56\linewidth}
  \centering
  \includegraphics[width=\linewidth]{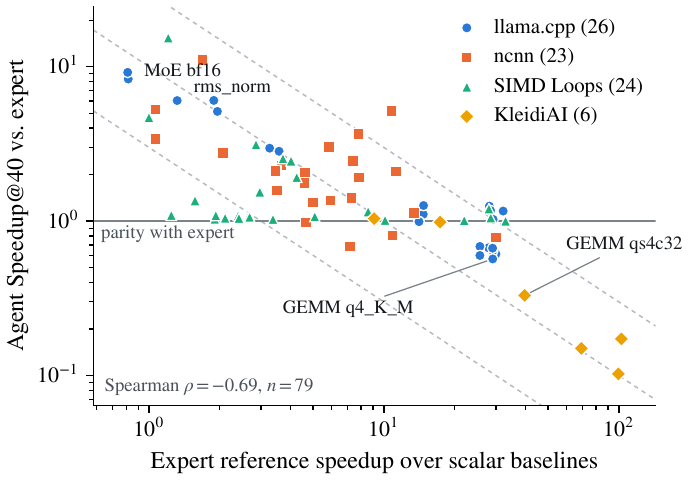}
\caption{\textbf{Agents gain more when expert kernels are less optimized.} One point per kernel.
$x$: expert speedup over the scalar reference; $y$: median Speedup@40 across
harness--model cells. Dashed lines show constant agent speedup over scalar (3$\times$, 10$\times$, 30$\times$).}
  \label{fig:where}
\end{minipage}\hfill
\begin{minipage}[t]{0.42\linewidth}
  \centering
  \includegraphics[width=\linewidth]{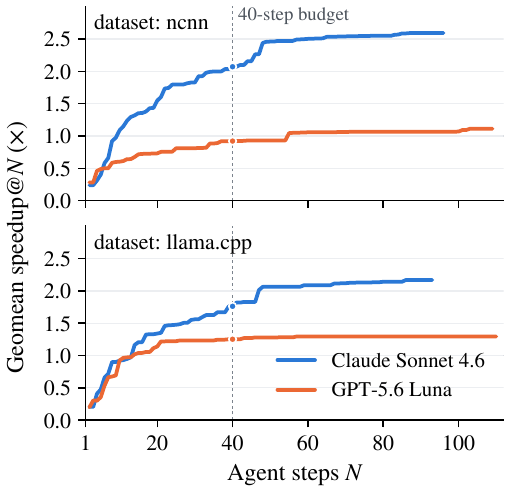}
  \caption{\textbf{Test-time scaling of agents.} Best-so-far Speedup@$t$ against the expert
  reference as a function of the tool-call budget $t$. Most agent progress is within 40 steps.}
  \label{fig:tts}
\end{minipage}
\end{figure}

Across the 79 SVE kernels, median agent Speedup@40 decreases as the expert
reference becomes stronger relative to unvectorized scalar code
(Spearman $\rho=-0.69$; $\rho=-0.75$ within llama.cpp; Figure~\ref{fig:where}).
Thus, large Speedup@40 values often reflect under-optimized reference paths
rather than unusually strong agent performance. This is clear in llama.cpp
RMSNorm: the reference is only $1.3\times$ faster than scalar, while agents
beat it by a median $6.0\times$. Conversely, for q4\_K\_M GEMM and MoE, the
expert implementations are already $25$--$29\times$ faster than scalar, and
agents reach only $0.57$--$0.69\times$ of the expert.

\subsection{What Improves Agent Optimization?}
\label{sec:improve}
\label{sec:rq4}
\begin{wraptable}{r}{0.42\linewidth}
  \vspace{-1.2em}
  \centering
  \caption{\textbf{Ablations} (nanobot + GPT-5.6-luna): SVE on the 32-kernel subset, \rev{SME2 on the
  20 KleidiAI SME2 kernels.}}
  \label{tab:ablation}
  \footnotesize
  \setlength{\tabcolsep}{4pt}
  \begin{tabular}{lc}
  \toprule
  Setting & Speedup@40 \\
  \midrule
  \rowcolor{gray!15}
  SVE & \\
  Control & 1.074 \\
  $+$ Arm docs & 0.997 \\
  $+$ cross-kernel notepad & 0.974 \\
  $-$ disassemble tool &  0.999 \\
  No intrinsics & 0.300 \\
  \midrule
  \rowcolor{gray!15}
  \rev{SME2, 20 KleidiAI kernels} & \\
  \rev{Control} & \rev{0.083} \\
  \rev{$+$ Arm docs} & \rev{0.130} \\
  \rev{$+$ SME2 pitfalls in prompt} & \rev{0.167} \\
  \bottomrule
  \end{tabular}
  \vspace{-1em}
\end{wraptable}
As shown in \autoref{fig:tts}, most of the improvement is realized within the first 40 tool calls,
but budget beyond 40 still pays off: Claude Sonnet 4.6 is $23\text{--}25\%$ higher at 100 calls than
at 40. This additional gain, however, comes mainly from a few kernels undergoing structural
rewrites between roughly 40 and 60 calls, and past 60 calls the curves are largely converged.

\hide{We judge the results in Table~\ref{tab:ablation} against the run-to-run variation, which has a standard deviation of $0.036$ (Section~\ref{sec:adapt}). Documentation modestly improves the agent's performance on SVE but hurts its performance on SME2. A notepad does not affect performance, and removing the disassemble tool does not hurt performance either, consistent with agents rarely using it. }

Arm documentation does not improve performance on SVE. On SME2 it raises Speedup@40 (\rev{$0.130$ vs.\ $0.083$}), but not through better SME2 code: \rev{with the documentation, no best version uses streaming SVE (vs.\ $6$ without it), $14$ of the $20$ use Neon (vs.\ $7$), and compile errors drop from $102$ to $54$.}
\rev{Putting common SME2 pitfalls in the prompt, with failure examples, doubles Speedup@40 on the 20
KleidiAI SME2 kernels ($0.083$ to $0.167$), but $5$ of the $20$ kernels get worse.}

Some tools contribute little in practice. The cross-kernel notepad does not improve performance, while removing the disassembly tool does not lower performance. Inspection of agent trajectories shows that agents rarely use disassembly, while human experts frequently use it while designing advanced kernels. In contrast, forbidding SIMD intrinsics, so that agents must rely on compiler auto-vectorization, lowers performance from $1.074$ to $0.300$: most of the agents' gains come from correctly using the provided intrinsics.
\label{sec:rq5}

\subsection{From Kernel Optimization to End-to-End Potential}
\label{sec:e2e}
\begin{wraptable}{r}{0.42\linewidth}
    \centering
    \caption{\textbf{End-to-end speedup from agent-optimized kernels.}
    Qwen3.5-4B and Llama-3.1-8B-Instruct (Q4\_K\_M) on Graviton4, relative to original llama.cpp.}
    \label{tab:e2e}
    \footnotesize
    \setlength{\tabcolsep}{3pt}
    \begin{tabular}{@{}lcccc@{}}
    \toprule
    & \multicolumn{2}{c}{\scriptsize Qwen3.5-4B} & \multicolumn{2}{c}{\scriptsize Llama-3.1-8B-Instruct} \\
    \cmidrule(lr){2-3}\cmidrule(l){4-5}
    Threads & Prefill & Decode & Prefill & Decode \\
    \midrule
    1  & 1.395 & 1.045 & 1.568 & 1.099 \\
    4  & 1.348 & 1.059 & 1.454 & 1.069 \\
    16 & 1.298 & 1.048 & 1.391 & 1.102 \\
    \bottomrule
    \end{tabular}
    \vspace{-1.5em}
  \end{wraptable}
  
To test whether kernel-level gains can translate into application-level improvements, we use Claude Code with Fable 5.1 to optimize 11 quantized matrix-multiplication kernels for Qwen3.5-4B and 7 kernels for Llama-3.1-8B-Instruct (both Q4\_K\_M), served with llama.cpp on Graviton4, and substitute the resulting shape-specialized implementations into the inference stack.

Compared with the original llama.cpp using its repacked i8mm fast path, the agent kernels speed up
prefill by $1.30$--$1.57\times$ and decode by $1.04$--$1.10\times$ across 1--16 threads for both
models (Table~\ref{tab:e2e}). Perplexity is unchanged on both models.

\paragraph{Gains come from adaptation to hardware.} All 18 kernels use Graviton4's i8mm instructions together with shape-specific tiling, layout, and prefetching. Individual kernels average 1.5$\times$ faster than llama.cpp's own repacked kernels if measured in isolation. Unlike llama.cpp's general kernels, the agent-generated kernels are specialized to the target models' shapes, so this shows the potential of automatic deployment-time specialization over expert libraries. Appendix~\ref{app:e2e} provides complete setup and detailed results.

%% file: tables/ladder-evolve.tex
\begin{table}[t]
\centering
\begin{minipage}[t]{0.5\linewidth}
\centering
\caption{\textbf{Instruction-set ladders on fixed hardware.} Holding the machine and task set
fixed, we progressively expose newer instructions; every arm is scored against the same reference. Speedup@40 of nanobot+GPT-5.6-luna, \rev{mean $\pm$ s.d.\ of 3 seeds}. Per-arm compile flags are in Appendix~\ref{app:isa-ladder}; failure analysis is in Appendix~\ref{app:m4-ladder-troubleshooting}.}
\label{tab:isa-ladder}
\footnotesize
\begin{tabular}{llc}
\toprule
Arm & Instructions available & Speedup@40 \\
\midrule
\rowcolor{gray!15}
\multicolumn{3}{l}{\emph{Graviton4, 32-kernel subset}} \\
L1 & Neon only                 & \rev{1.105\,{\scriptsize$\pm$.07}} \\
L2 & $+$ SVE                   & \rev{0.953\,{\scriptsize$\pm$.06}} \\
L3 & $+$ SVE2                  & \rev{1.101\,{\scriptsize$\pm$.01}} \\
\midrule
\rowcolor{gray!15}
\multicolumn{3}{l}{\rev{\emph{Apple M4, KleidiAI SME2 set (20 kernels)}}} \\
L4 & Neon only                 &  \rev{0.141\,{\scriptsize$\pm$.01}} \\
L5 & $+$ streaming SVE (no ZA) &  \rev{0.059\,{\scriptsize$\pm$.004}} \\
L6 & $+$ ZA / \texttt{FMOPA}   &  \rev{0.085\,{\scriptsize$\pm$.006}} \\
\bottomrule
\end{tabular}
\end{minipage}\hfill
\begin{minipage}[t]{0.47\linewidth}
\centering
\caption{\textbf{Cross-generation adaptation.} Speedup@40
against the expert implementation of the older (Before) and newer (After) generation;
Ratio = After / Before. \rev{Mean $\pm$ s.d.\ of 3 seeds.}}
\label{tab:evolve}
\footnotesize
\setlength{\tabcolsep}{3.5pt}
\begin{tabular}{lccc}
\toprule
Model & Before & After & Ratio \\
\midrule
\rowcolor{gray!15}
\multicolumn{4}{l}{\rev{\emph{KleidiAI: SVE (6) $\to$ SME2 (6)}}} \\
Claude Opus 4.8 & \rev{0.436\,{\scriptsize$\pm$.02}} & \rev{0.562\,{\scriptsize$\pm$.20}} & \rev{1.29} \\
Claude Sonnet 4.6 & \rev{0.258\,{\scriptsize$\pm$.02}} & \rev{0.266\,{\scriptsize$\pm$.02}} & \rev{1.03} \\
DeepSeek-V4-Pro & \rev{0.213\,{\scriptsize$\pm$.03}} & \rev{0.091\,{\scriptsize$\pm$.01}} & \rev{0.42} \\
GLM-5.3 & \rev{0.464\,{\scriptsize$\pm$.05}} & \rev{0.213\,{\scriptsize$\pm$.19}} & \rev{0.46} \\
\rev{GPT-5.6-luna} & \rev{0.178\,{\scriptsize$\pm$.04}} & \rev{0.073\,{\scriptsize$\pm$.01}} & \rev{0.41} \\
\midrule
\rowcolor{gray!15}
\multicolumn{4}{l}{\emph{SIMD Loops: SVE (8) $\to$ SVE2 (8)}} \\
Claude Opus 4.8   & 1.859\,{\scriptsize$\pm$.21} & 1.957\,{\scriptsize$\pm$.02} & 1.05 \\
Claude Sonnet 4.6 & 1.750\,{\scriptsize$\pm$.18} & 1.752\,{\scriptsize$\pm$.12} & 1.00 \\
DeepSeek-V4-Pro   & 1.587\,{\scriptsize$\pm$.18} & 1.504\,{\scriptsize$\pm$.16} & 0.95 \\
GLM-5.3           & 1.929\,{\scriptsize$\pm$.07} & 1.810\,{\scriptsize$\pm$.19} & 0.94 \\
\bottomrule
\end{tabular}
\end{minipage}
\end{table}

%% file: sections/02_related_work.tex
\section{Related Work}
\label{sec:related}

\paragraph{Kernel optimization across hardware.}
Recent work evaluates whether language models and coding agents can generate efficient kernels
for a given hardware target or programming interface. KernelBench and TritonBench study GPU
kernel generation~\citep{ouyang2025kernelbench,li2025tritonbench}, while MobileKernelBench
and SimdBench extend to mobile CPUs and SIMD intrinsics%
~\citep{zou2026mobilekernelbench,he2025simdbench}.
Other benchmarks extend kernel evaluation across architectures and devices%
~\citep{li2026codegenbench,wen2025multikernelbench,zang2026kernelgenbench,chen2025eok}. These works primarily evaluate performance on a fixed target or a set of
hardware targets. In contrast, \bench{} organizes hardware as a sequence of successive
architectural extensions and evaluates matched kernel tasks across generations where
coverage permits. This lets us study adaptation to new instructions and programming
abstractions.

\hide{A growing body of work evaluates LLMs for generating performance-critical kernels.
On GPUs, KernelBench, TritonBench, and SOL-ExecBench evaluate generated CUDA or Triton
kernels~\citep{ouyang2025kernelbench,li2025tritonbench,lin2026solexecbench},
while GPU Kernel Scientist, CudaForge, and CUDA Agent introduce iterative or agentic
optimization with hardware feedback%
~\citep{andrews2025gpukernelscientist,zhang2025cudaforge,dai2026cudaagent}.
CPU-oriented work studies mobile inference kernels%
~\citep{zou2026mobilekernelbench},
cross-architecture HPC kernels%
~\citep{li2026codegenbench},
SIMD-intrinsic generation%
~\citep{he2025simdbench,li2026autoveccoder},
and optimization in emerging RISC-V environments%
~\citep{chen2025eok}.
Recent benchmarks broaden kernel evaluation across hardware:
MultiKernelBench supports heterogeneous GPU, NPU, and TPU backends,
while KernelGenBench evaluates LLM- and agent-generated kernels across multiple chips%
~\citep{wen2025multikernelbench,zang2026kernelgenbench}.
These works treat hardware platforms mainly as parallel targets.
In contrast, \bench{} follows successive generations within one architecture family
(SVE$\to$SVE2$\to$SME2), using hardware evolution itself to test whether agent systems
can adapt to newly introduced programming abstractions.}

\paragraph{Coding agents for performance optimization.}
Recent coding agents improve programs through iterative execution and feedback~\citep{yang2024sweagent,wang2025openhands}.
For kernel optimization, GPU Kernel Scientist and CudaForge use iterative search and hardware
feedback, while AdaExplore combines execution-driven adaptation with
diversity-preserving search%
~\citep{andrews2025gpukernelscientist,zhang2025cudaforge,du2026adaexplore}.
More recent systems train kernel-optimization agents
with reinforcement learning%
~\citep{dai2026cudaagent,fu2026davincikernel}.
These works show rapid progress in agentic kernel optimization.
\bench{} provides both a controlled benchmark and a hardware-in-the-loop environment for
testing whether such agents can adapt their optimization strategies as hardware evolves.

\hide{Coding models have been evaluated from function-level code generation%
~\citep{chen2021codex,austin2021mbpp,jain2025livecodebench}
to repository-level software engineering with interactive agents%
~\citep{jimenez2024swebench,yang2024sweagent,wang2025openhands}.
Recent benchmarks further study performance engineering on real repositories and workloads,
where agents must produce correctness-preserving optimizations and are often compared
against expert patches%
~\citep{he2026sweperf,ma2026swefficiency,garg2025perfbench,cui2026perfoptbench}.
These evaluations largely operate on fixed repository snapshots, dependencies, and execution
environments, testing whether agents can solve tasks drawn from an existing software distribution.}

\paragraph{Automated program optimization.}
A broader line of work studies program optimization using learned cost models,
reinforcement learning, and search%
~\citep{chen2018tvm,zheng2020ansor,trofin2021mlgo,cummins2022compilergym,
wei2025supercoder,mankowitz2023alphadev,fawzi2022alphatensor,novikov2025alphaevolve}.
However, automated optimizers can still fall short of expert implementations when performance
depends on hardware-specific transformations%
~\citep{mendis2019revec,tschand2026jaxbench}.
This gap motivates coding agents that can optimize kernels through interaction
with real hardware.

\hide{More broadly, learned cost models, reinforcement learning, and LLM-guided search
have been applied to tensor scheduling, compiler optimization, superoptimization,
and algorithm discovery%
~\citep{chen2018tvm,zheng2020ansor,trofin2021mlgo,cummins2022compilergym,
wei2025supercoder,mankowitz2023alphadev,fawzi2022alphatensor,novikov2025alphaevolve}.
These works develop optimization methods, often around compiler representations,
structured search spaces, or specialized search loops.
In contrast, our focus is evaluation: we study how general-purpose coding agent systems
optimize free-form kernels and adapt as the target hardware itself evolves.}

%% file: sections/07_conclusion.tex
\section{Conclusion}
\label{sec:conclusion}

We introduced \bench{}, a benchmarking pipeline and set of environments to study how well coding agents optimize kernels as hardware evolves. Our results show that stronger agents can match or beat production kernels on established vector ISAs, but they struggle with programming-model shifts such as SME2. At the same time, the end-to-end speedups that agent-written kernels deliver on two LLMs highlight the potential of agent-driven kernel optimization in real deployments.
\paragraph{Limitations.}
Newer hardware is harder to evaluate because strong expert implementations are often not available, which limits how many kernels we can compare across generations; in particular, our SME2 evidence rests on the 20 KleidiAI kernels that have an SME2 expert reference. Our experiments also focus on Arm CPUs, so it remains to be seen whether the same trends hold for other architectures such as x86 or GPUs. Finally, our kernel-level evaluation is single-core and does not cover all effects that may appear in multi-core deployment.

%% file: sections/99_statements.tex
\subsection*{AI Use Statement}
AI agents are the primary subject of this study. We also used AI tools to assist with research code generation, benchmark data collection and synthesis, and wording refinement. All AI-generated code, benchmark instances, and paper refinements were reviewed and verified by the authors. The authors were responsible for the code, experiments, and manuscript.

\subsection*{Ethics Statement}
\bench{} is constructed from open-source software, and all redistributed code follows the licenses of the original repositories: ncnn (BSD-3-Clause), llama.cpp (MIT), Arm SIMD Loops (BSD-3-Clause), and KleidiAI (Apache-2.0). %
The study does not involve human subjects.

\subsection*{Reproducibility Statement}

\ifpreprint
The code for the benchmark, including the scripts for evaluating the coding agents, is available at \url{https://github.com/cmu-l3/kernel-shift}, and the dataset at \url{https://huggingface.co/datasets/l3lab/KernelShift}.
\fi

\ifpreprint
\subsection*{Acknowledgments}
This work was supported in part by SoftBank Group Corp. and Arm. This program was made possible (in part) due to the generosity of SoftBank Group Corp. We thank our collaborators at Arm for their support and collaboration.
\fi

%% file: sections/appendix.tex
\section{Benchmark Composition Details}
\label{app:composition}

\subsection{Composition by source}
\bench{} contains 99 kernel specifications in the evaluation set across four source families:
23 from ncnn~\citep{ncnn}, 26 from llama.cpp~\citep{gerganov2023llamacpp}, 24 from Arm SIMD
Loops~\citep{armsimdloops}, and 26 from KleidiAI~\citep{kleidiai}.
The KleidiAI inventory comprises six SVE tasks and 20 SME2 tasks.

Table~\ref{tab:sources} lists the operators each source contributes, their precisions, and where
the workload shapes and the expert reference come from.

\begin{table}[!ht]
\caption{\textbf{Kernel sources in \bench{}.} Numbers in parentheses are kernel counts.}
\label{tab:sources}
\centering\footnotesize
\setlength{\tabcolsep}{4pt}
\begin{tabular}{@{}>{\raggedright\arraybackslash}p{1.35cm}>{\raggedright\arraybackslash}p{3.9cm}>{\raggedright\arraybackslash}p{1.55cm}>{\raggedright\arraybackslash}p{2.95cm}>{\raggedright\arraybackslash}p{1.95cm}r@{}}
\toprule
Source & Kernels & Precisions & Workload shapes & Expert reference & \# \\
\midrule
ncnn
  & conv2d (10), depthwise conv2d (8), pooling (5)
  & FP32, INT8 (w8a8ch)
  & 6 vision models: ResNet50, MobileNetV3-Large, VGG16, GoogLeNet, AlexNet, SqueezeNet1.1
  & ncnn Arm (Neon) layers
  & 23 \\
\cmidrule{1-6}
llama.cpp
  & GEMM (15), MoE (6), attention: MHA, GQA, MLA prefill, MLA decode (4), RMSNorm (1)
  & BF16, q8\_0, q4\_K\_M, FP32
  & 5 LLMs on ShareGPT prompts: Qwen1.5-MoE-A2.7B, OLMoE-1B-7B, DeepSeek-V3, Llama-3.1-8B, Mistral-7B
  & ggml CPU kernels
  & 26 \\
\cmidrule{1-6}
Arm SIMD Loops
  & reductions and dot products (9), element-wise arithmetic (4), complex arithmetic (4),
    convolution and correlation (2), string and pixel processing (2),
    search and conditional update (2), INT8 matrix--vector product (1)
  & FP16--FP64, INT8--INT64
  & edge-case and performance sizes
  & Arm's hand-written SVE version of each loop
  & 24 \\
\cmidrule{1-6}
KleidiAI
  & \rev{SVE: GEMM (5; FP32, INT8, INT4), conv2d $3{\times}3$ (1).
    SME2: GEMM (16; FP32, FP16, BF16, INT8, INT4, INT2), conv2d $3{\times}3$ (2), depthwise conv2d $3{\times}3$ (2)}
  & \rev{FP32, FP16, BF16, INT8, INT4, INT2}
  & edge-case and performance sizes
  & KleidiAI SVE and SME2 micro-kernels
  & \rev{6 + 20} \\
\midrule
Total & & & & & 99 \\
\bottomrule
\end{tabular}
\end{table}

\subsection{Task sets per experiment}
\label{app:coverage}
Table~\ref{tab:sources} lists what the benchmark contains; Table~\ref{tab:task-sets} lists which
kernels each experiment uses, on which machine, and against which reference. The main
comparison uses the full SVE set, and the other experiments use subsets chosen so that every arm or
generation being compared has the same kernels and valid reference baselines.

\begin{table}[!ht]
\caption{\textbf{Task set of each experiment.} Kernel counts are ncnn\,/\,llama.cpp\,/\,SIMD Loops\,/\,KleidiAI. G3, G4: Graviton3, Graviton4.}
\label{tab:task-sets}
\centering\footnotesize
\setlength{\tabcolsep}{3pt}
\begin{tabular}{@{}>{\raggedright\arraybackslash}p{2.5cm}ll>{\raggedright\arraybackslash}p{2.75cm}>{\raggedright\arraybackslash}p{2.6cm}>{\raggedright\arraybackslash}p{2.45cm}@{}}
\toprule
Experiment & Where & Machine & Harness, model & Kernels & Reference \\
\midrule
Main comparison & Tab.~\ref{tab:main-sve} & G3 & all 14 pairs & 23\,/\,26\,/\,24\,/\,6 (79) & expert \\
\cmidrule{1-6}
ISA ladder L1--L3 & Tab.~\ref{tab:isa-ladder} & G4 & nanobot, GPT-5.6-luna & 32-subset: 10\,/\,9\,/\,12\,/\,1 & expert, same in every arm \\
ISA ladder L4--L6 & Tab.~\ref{tab:isa-ladder} & M4 & nanobot, GPT-5.6-luna & KleidiAI SME2: 20 & expert, same in every arm \\
\cmidrule{1-6}
KleidiAI SVE$\to$SME2 & Tab.~\ref{tab:evolve} & G3$\to$M4 & nanobot, 5 models & KleidiAI: 6 $\to$ 6 & KleidiAI SVE $\to$ SME2 \\
SIMD Loops SVE$\to$SVE2 & Tab.~\ref{tab:evolve} & G3$\to$G4 & nanobot, 4 models & SIMD Loops: 8 & Arm SVE $\to$ SVE2 versions \\
\cmidrule{1-6}
Ablations, SVE & Tab.~\ref{tab:ablation} & G3 & nanobot, GPT-5.6-luna & 32-subset: 10\,/\,9\,/\,12\,/\,1 & expert \\
Ablations, SME2 & Tab.~\ref{tab:ablation} & M4 & nanobot, GPT-5.6-luna & 32-subset (SME2 version): 10\,/\,9\,/\,12\,/\,1 & expert \\
Test-time scaling & Fig.~\ref{fig:tts} & G3 & nanobot, GPT-5.6-luna; Claude Sonnet 4.6 & 11\,/\,8\,/\,0\,/\,0 (19) & expert \\
\cmidrule{1-6}
End-to-end & App.~\ref{app:e2e} & G4 & Claude Code, Claude Fable 5.1 & 11 Qwen3.5-4B and 7 Llama-3.1-8B-Instruct GEMMs & ggml; original llama.cpp \\
\bottomrule
\end{tabular}
\end{table}

\paragraph{32-kernel subset.}
A dtype-balanced sample of the benchmark. 
FP32 and w8a8ch versions of depthwise $3{\times}3$ and $5{\times}5$ (stride 1),
conv2d $3{\times}3$ (stride 1) and $7{\times}7$ (stride 2), global average pooling, and $3{\times}3$
max pooling (stride 2). llama.cpp: GEMM BF16, q4\_K\_M and q8\_0 ($N{=}K{=}2048$), GEMM q8\_0
($N{=}1408$), MHA, and MoE BF16, q4\_K\_M and q8\_0 (OLMoE). SIMD Loops: \texttt{loop\_001},
\texttt{002}, \texttt{003}, \texttt{024}, \texttt{027}, \texttt{035}, \texttt{037},
\texttt{105}, \texttt{108}, \texttt{113}, \texttt{126}, \texttt{127}. KleidiAI: FP32 GEMM, in its SVE
version on Graviton and its SME2 version on M4.

\paragraph{SIMD Loops across generations.}
\texttt{loop\_103}, \texttt{105}, \texttt{108}, \texttt{109}, \texttt{110}, \texttt{112},
\texttt{113}, and \texttt{114}: loops whose Arm SVE2 implementation does not compile without
SVE2 (i.e., it uses SVE2-only instructions such as \texttt{svcdot}, \texttt{svcmla}, or
\texttt{svmatch}). 
Table~\ref{tab:evolve} reports the mean of three seeds.

\paragraph{Test-time scaling subset.}
19 ncnn and llama.cpp kernels, mostly from the 32-kernel subset. The Claude Sonnet 4.6 curves use
the main-comparison runs restricted to these kernels.

\paragraph{End-to-end kernels.}
The 11 quantized GEMM shapes of Qwen3.5-4B, given as $(N, K)$: q4\_K at (1024, 2560), (2560, 4096),
(2560, 9216), (4096, 2560), (8192, 2560), and (9216, 2560); q5\_K at (2560, 4096) and (8192, 2560);
q6\_K at (1024, 2560), (248320, 2560), and (2560, 9216); and 7 quantized GEMM shapes of Llama-3.1-8B-Instruct, given as $(N,K)$: q4\_K at (1024, 4096), (4096, 4096), (4096, 14336), and (14336,
  4096); q6\_K at (1024, 4096), (4096, 14336), and (128256, 4096).

\subsection{Kernel inventory}
\label{app:kernel-list}
Most kernels in \bench{} are instances of a few operator families that vary in precision and
in fixed operator attributes. Table~\ref{tab:kernel-list} lists all 99 kernels in this compact
form: each row is an operator family, and its variants expand to the listed number of kernels.
Unless stated otherwise, convolution variants are written as kernel size and stride, GEMM
variants as $(N, K)$, and every kernel runs on SVE.

\begin{table}[!ht]
\caption{\textbf{All \rev{99} kernels by operator family.} $\times$: cross product; s: stride; SIMD Loops are given by loop number. Qwen: Qwen1.5-MoE-A2.7B; OLMoE: OLMoE-1B-7B.}
\label{tab:kernel-list}
\centering\footnotesize
\setlength{\tabcolsep}{4pt}
\begin{tabular}{@{}l>{\raggedright\arraybackslash}p{3.3cm}>{\raggedright\arraybackslash}p{7.8cm}r@{}}
\toprule
Source & Family & Variants & \# \\
\midrule
ncnn & conv2d & \{FP32, w8a8ch\} $\times$ \{$1{\times}1$ s1, $1{\times}1$ s2, $3{\times}3$ s1, $3{\times}3$ s2, $7{\times}7$ s2\} & 10 \\
     & depthwise conv2d & \{FP32, w8a8ch\} $\times$ \{$3{\times}3$, $5{\times}5$\} $\times$ \{s1, s2\} & 8 \\
     & pooling (FP32) & global average; max $2{\times}2$ s2; max $3{\times}3$ s1; max $3{\times}3$ s2 with and without padding & 5 \\
\cmidrule{1-4}
llama.cpp & GEMM & \{BF16, q8\_0, q4\_K\_M\} $\times$ \{(1024, 2048), (1408, 2048), (2048, 1024), (2048, 1408), (2048, 2048)\}; q4\_K\_M uses $K{=}1536$ in place of 1408 & 15 \\
          & MoE & \{BF16, q8\_0, q4\_K\_M\} $\times$ \{Qwen: 60 experts, top-4; OLMoE: 64 experts, top-8\} & 6 \\
          & attention (BF16) & MHA (Qwen); GQA, 32 query and 8 KV heads (Llama-3.1-8B, Mistral-7B); MLA prefill and MLA decode (DeepSeek-V3) & 4 \\
          & RMSNorm (FP32) & hidden size 2048 (Qwen) & 1 \\
\cmidrule{1-4}
SIMD Loops & reductions, dot products & \texttt{001}, \texttt{002}, \texttt{003}, \texttt{004}, \texttt{008}, \texttt{024}, \texttt{033}, \texttt{105}, \texttt{113} & 9 \\
           & element-wise arithmetic & \texttt{027}, \texttt{028}, \texttt{029}, \texttt{035} & 4 \\
           & complex arithmetic & \texttt{037}, \texttt{109}, \texttt{110}, \texttt{112} & 4 \\
           & convolution, correlation & \texttt{038}, \texttt{114} & 2 \\
           & string, pixel processing & \texttt{103}, \texttt{108} & 2 \\
           & search, conditional & \texttt{126}, \texttt{127} & 2 \\
           & matrix--vector & \texttt{219} (INT8) & 1 \\
\cmidrule{1-4}
KleidiAI & GEMM, SVE & \rev{FP32 (512, 512); INT8 and INT4 weights (512, 512); INT4 weights with INT8 activations, FP16 or BF16 block scales (256, 512)} & \rev{5} \\
         & conv2d, SVE & FP32 NHWC $3{\times}3$ s1, 32$\to$64 channels & 1 \\
         & GEMM, SME2 & \rev{\{FP32, FP16, BF16\} (512, 512); at (256, 512): INT8, INT4, and INT2 weights with FP32 activations (7), INT8 and INT4 weights with FP16 activations (4), UINT8 inputs (2)} & \rev{16} \\
         & conv2d, SME2 & \rev{\{FP16 valid, FP32 s1\} NHWC $3{\times}3$, 32$\to$64 channels} & \rev{2} \\
         & depthwise conv2d, SME2 & \{FP32, FP16\} NHWC $3{\times}3$ s1 & 2 \\
\midrule
Total & & & \rev{99} \\
\bottomrule
\end{tabular}
\end{table}

\paragraph{Inclusion criteria.}
A constructed kernel enters the benchmark only if it has a usable expert reference to measure the agent's relative performance. After a
task is constructed, we build and run the expert implementation on the target machine and flag
the kernel if (i) no expert implementation could be obtained from the source, (ii) the expert
implementation fails to build or does not pass the correctness check on the task's
workloads, or (iii) it is 2x slower than the compiler auto-vectorized version of the scalar starting
point, in which case it does not represent expert performance. Flagged kernels are inspected
and removed, and the remaining kernels form the inventory in Table~\ref{tab:kernel-list}.
\rev{The released dataset contains 144 kernel definitions, containing the 99 evaluated kernels and the 18 GEMMs of the
end-to-end study (Section~\ref{sec:e2e}).}

\hide{
\subsection{Source assets and licenses}
\label{app:licenses}
\todo{\bench{} redistributes code derived from four open-source repositories, each under a
permissive license: ncnn~\citep{ncnn} (BSD-3-Clause; some third-party components under other
terms), llama.cpp~\citep{gerganov2023llamacpp} (MIT), Arm SIMD Loops~\citep{armsimdloops}
(BSD-3-Clause), and KleidiAI~\citep{kleidiai} (Apache-2.0).
\begin{center}\small
\begin{tabular}{lll}
\toprule
Asset & Use in \bench{} & License \\
\midrule
ncnn            & scalar/expert kernels, bindings & BSD-3-Clause \\
llama.cpp (ggml) & scalar/expert kernels, shape tracing & MIT \\
Arm SIMD Loops  & scalar and SVE kernels          & BSD-3-Clause \\
KleidiAI        & expert micro-kernels            & Apache-2.0 \\
\midrule
ShareGPT~\citep{sharegpt} & prompts for shape tracing only & Apache-2.0 \\
\bottomrule
\end{tabular}
\end{center}
\note{Also list the assets used only to derive shapes: torchvision models (BSD-3-Clause),
the Hugging Face configs and GGUF checkpoints traced through llama.cpp (Qwen1.5-MoE, OLMoE,
DeepSeek-V3, Llama-3.1-8B, Mistral-7B; each under its own model license).}}}

\section{Task Construction}
\label{sec:app-construction}

\subsection{Construction skills and validation scripts}
\label{app:skills}
Task construction is carried out by a coding agent that follows written procedures, which we
call \emph{construction skills}. Each skill is a document that the agent loads on demand. It
states what to extract from the request and the given resources, the artifact to produce and its schema, the checks to run afterwards, and a list of known failure modes.
Each skill calls deterministic scripts for the mechanical steps, such as capturing workload shapes
and building and running a task. The agent supplies the judgment these scripts cannot, such as
which dimensions are fixed per kernel, how to call the source library, and which workloads fill
a coverage gap, and it runs the validation on the target machine. For the entire generation pipeline, we use Claude Code with sonnet-5 as the base model.

Four skills cover the benchmarking pipeline of Section~\ref{sec:pipeline}. \texttt{gen-definition} writes the
kernel specification (Figure~\ref{fig:skill-gen-definition}), \texttt{gen-workload} collects its
workloads (Figure~\ref{fig:skill-gen-workload}), \texttt{gen-candidate-solution} renders the scalar
starting implementation (Figure~\ref{fig:skill-gen-candidate}), and \texttt{gen-baseline-solution}
wraps the source library's kernel as the expert reference (Figure~\ref{fig:skill-gen-baseline}).
The figures show condensed versions; the complete skill documents are released with the
benchmark. The skills stop for human input only at a few fixed points, described next.

\paragraph{Human involvement.}
Humans make three kinds of decisions. They select kernels, approving each definition that
\texttt{gen-definition} proposes and, for large libraries such as KleidiAI with hundreds of
micro-kernels, choosing a representative subset. They approve the agent-proposed workloads of Arm SIMD Loops and KleidiAI
(Appendix~\ref{app:workloads}). They review each expert reference after its first evaluation, since a correct wrapper
can still underperform the library. If such a wrapper cannot be fixed, the kernel is excluded
(Appendix~\ref{app:kernel-list}). The scalar starting implementations need no human input.
Appendix~\ref{app:extension-cost} shows what humans did in one extension.

\begin{figure}[!ht]
\begin{AIbox}{Construction Skill: \texttt{gen-definition}}
\small
\textbf{Task Description}

Design and write a Definition JSON for a new operator. This skill writes the structural
schema only; concrete tensor sizes are added afterwards by \texttt{gen-workload}. The
reference function inside the definition is a correctness oracle, not a workload.

\textbf{What to Extract from the Request}

\verb|  |* Op name and what it computes (enough to write a reference function)

\verb|  |* Axes: which are fixed per kernel configuration (\texttt{const}) and which vary per workload (\texttt{var})

\verb|  |* Input and output tensors: names, shapes (as lists of axis names), and dtypes

\verb|  |* Constraints expressing derived output axes in terms of other axes

\verb|  |* A reference implementation in NumPy or PyTorch

If the request is underspecified in any of these areas, ask before generating.

\textbf{Workflow}

\verb|  |0. If the op comes from a real model, extract its shapes from the Hugging Face \texttt{config.json} or from torchvision forward hooks (no weights are downloaded), and record the source as a \texttt{model:<name>} tag.

\verb|  |1. Read one or two existing definitions of the closest operator family.

\verb|  |2. Elicit the design from the user until it is unambiguous, including which framework supplies the expert reference (\texttt{baseline-solution:<framework>} tag). \textbf{Always ask; do not infer it from the model or the dtype.}

\verb|  |3. Present the complete definition and its output path. \textbf{Wait for explicit confirmation before writing.}

\verb|  |4. Write the JSON, then check that every axis used in a shape or constraint is declared, every \texttt{const} axis has an integer value, every derived \texttt{var} axis has a constraint, and the reference function uses integer literals rather than axis names.

\verb|  |5. Verify that the benchmark loads the definition; diagnose and fix any error before declaring success.

[\ldots{} conventions and known failure modes omitted \ldots]
\end{AIbox}
\caption{\textbf{Construction skill for kernel definitions}, condensed from the skill document
given to the construction agent (about 1{,}900 words). Command lines, file paths, and the table of
known failure modes are omitted.}
\label{fig:skill-gen-definition}
\end{figure}

\begin{figure}[!ht]
\begin{AIbox}{Construction Skill: \texttt{gen-workload}}
\small
\textbf{Task Description}

Add workloads to the definitions of an operator family. The primary path is always automated
collection; manual additions are for targeted gap-fills only.

\textbf{Workflow}

\verb|  |1. Identify the operator family and its collection path:

\verb|    |* Vision kernels: hook the matching module of torchvision ResNet50 and MobileNetV3-Large over 11 input resolutions (96 to 1024, plus 360$\times$640 and 480$\times$640).

\verb|    |* LLM kernels: run llama.cpp with a GGUF model on 200 ShareGPT prompts on a Graviton instance and record the operator shapes.

\verb|    |* Neither: go to manual addition (step 4).

Captured shapes are matched to the definition by its own \texttt{const} axes, so a new definition in a supported family needs no script change.

\verb|  |2. Run the collector, previewing with a dry run first.

\verb|  |3. The collected pool is filtered automatically: keep at most two copies of any identical axis combination; seed one workload per distinct value of a batch axis (\texttt{C\_in} for convolution, \texttt{C} for depthwise and pooling, \texttt{M} for GEMM, \texttt{n\_tokens} for MoE); fill the remaining slots greedily by farthest-point selection in normalized axis space, up to 20 workloads.

\verb|  |4. For manual additions, \textbf{present a table of the planned workloads and their rationale, and wait for explicit confirmation before writing.} Check coverage: at least one channel count that is not a multiple of common SIMD widths; 15--20\% of workloads with an odd spatial dimension; for attention, the decode point $M{=}1$ at the maximum context length.

\verb|  |5. Verify the workload count and that the benchmark loads them.

[\ldots{} conventions, command lines, per-family axis tables, and known failure modes omitted \ldots]
\end{AIbox}
\caption{\textbf{Construction skill for workloads}, condensed from the skill document (about
1{,}500 words).}
\label{fig:skill-gen-workload}
\end{figure}

\begin{figure}[!ht]
\begin{AIbox}{Construction Skill: \texttt{gen-candidate-solution}}
\small
\textbf{Task Description}

Generate the scalar starting implementation for the definitions of an operator. Each solution
embeds three sources: a header holding the harness contract with the definition's
\texttt{const} axes baked in as \texttt{constexpr}, a binding that implements the common entry
symbol \texttt{kernelshift\_entry\_<op\_type>}, and a kernel file with plain scalar loops, which is
the file the optimizing agent later replaces. Everything is derived from the definition; there is
no configuration file.

\textbf{What to Extract from the Request}

\verb|  |* Op type, and whether to target one definition or all of them

\verb|  |* Variant: plain FP32 or a quantized scheme, read from the definition's input dtypes

\verb|  |* Mode: a new operator or variant, or regeneration of existing solutions

\textbf{Workflow}

\verb|  |0. Check the definition's \texttt{baseline-solution} tag and generate only for definitions of the matching source.

\verb|  |1. Classify the definition: \texttt{const} axes become \texttt{constexpr}; \texttt{var} axes become entry parameters; a scalar input is baked only if it has the same value in every workload of the definition, which is checked against the workload file, and the definition is skipped otherwise.

\verb|  |2. Derive the entry signature from the definition: tensor pointers in declaration order with the output second, each typed by its declared dtype, followed by the \texttt{var} dimensions deduplicated across all tensor shapes in first-seen order.

\verb|  |3. Write the header, binding, and scalar kernel templates. For a quantized kernel, match the definition's Python reference: accumulate in 64-bit integers, then apply the dequantization scales and bias in double precision.

\verb|  |4. Generate the solutions (dry run first), then compile and run each on every workload on the target machine.

[\ldots{} rules, command lines, template file layout, and known failure modes omitted \ldots]
\end{AIbox}
\caption{\textbf{Construction skill for the scalar starting implementation}, condensed from the
skill document (about 1{,}800 words).}
\label{fig:skill-gen-candidate}
\end{figure}

\begin{figure}[!ht]
\begin{AIbox}{Construction Skill: \texttt{gen-baseline-solution}}
\small
\textbf{Task Description}

Generate the expert-reference solution for the definitions of an operator by wrapping the
source library's kernel. The solution has the same three sources as the starting
implementation, but its kernel file delegates to the library (for ncnn, the Arm layer classes
linked from the static library). One generator serves every source library; a new library is
described by parameters, not by a new script.

\textbf{What to Extract from the Request}

\verb|  |* Op type and scope (one definition or all definitions tagged for this library)

\verb|  |* Library: if it is not an existing one, this is new-library setup; \textbf{confirm the scope with the user}

\verb|  |* Mode: a new operator or regeneration after the definitions changed

\textbf{Workflow}

\verb|  |1. Nothing needs registering: definitions are discovered from the definition directory, and every template value is derived from the definition itself.

\verb|  |2. Learn the library API for this operator by \textbf{reading the library's real headers rather than assuming it from another operator}.

\verb|  |3. Write the header, binding, and kernel templates. Bake only \texttt{const} axes and workload-invariant scalars; derive every \texttt{var} dimension at run time from the tensor shapes. A template with an unresolved placeholder is the wrong template for that definition, and the definition is skipped.

\verb|  |4. Generate the solutions, verify that the benchmark loads them, then collect the baselines on the target machine to confirm that they compile and pass correctness.

[\ldots{} rules, command lines, per-layer API notes, and known failure modes omitted \ldots]
\end{AIbox}
\caption{\textbf{Construction skill for the expert reference}, condensed from the skill document
(about 2{,}800 words). The omitted per-layer notes document initialization pitfalls
of individual ncnn layer classes.}
\label{fig:skill-gen-baseline}
\end{figure}

\subsection{Workloads}
\label{app:workloads}
Workloads for ncnn and llama.cpp kernels come from model executions: in ncnn, operator shapes observed
in ResNet50 and MobileNetV3-Large at 11 input resolutions, and in llama.cpp inference on 200
ShareGPT prompts~\citep{sharegpt} of 10--1500 words with a context of up to 4096 tokens. The
collected shapes are deduplicated and subsampled to about 20 per kernel
(Figure~\ref{fig:skill-gen-workload}). Arm SIMD Loops and KleidiAI have no model execution to
trace; their workloads are proposed by the agent and approved by humans, and combine small
edge-case sizes for correctness with large sizes for performance.

Only shapes are kept from the executions. Input tensors are generated deterministically based on
each workload's ID, and quantization scales are drawn from fixed ranges, so every evaluation of a
workload sees the same inputs. %
Across the 99 kernels, the benchmark has 1{,}582 workloads.

\subsection{Correctness criteria}
\label{app:correctness}
An output element matches if it is within both an absolute and a relative tolerance
of the reference value; positions where the reference itself is not finite are ignored. A
workload passes if the fraction of matching elements reaches the required match ratio, and a
candidate is correct only if it passes every workload. Kernels whose error is dominated by
quantization or low-precision accumulation are instead checked with the signal-to-quantization-noise
ratio, $\mathrm{SQNR}=10\log_{10}\big(\sum_i r_i^2/\sum_i(y_i-r_i)^2\big)$, against a minimum.
Table~\ref{tab:tolerances} lists the thresholds.

\begin{table}[!ht]
\caption{\textbf{Correctness thresholds.} Kernels not listed use the default.}
\label{tab:tolerances}
\centering\footnotesize
\begin{tabular}{@{}lccc@{}}
\toprule
Kernels & Absolute tol. & Relative tol. & Match ratio \\
\midrule
Default (convolution, pooling, RMSNorm, SIMD Loops) & $10^{-3}$ & $10^{-3}$ & 1.00 \\
GEMM & $2\times10^{-3}$ & $10^{-2}$ & 0.98 \\
MoE & $10^{-2}$ & $5\times10^{-2}$ & 0.95 \\
MHA & $10^{-3}$ & $10^{-3}$ & 0.98 \\
GQA, MLA prefill & $5\times10^{-3}$ & $10^{-3}$ & 0.98 \\
MLA decode & $3\times10^{-2}$ & $10^{-2}$ & 0.98 \\
KleidiAI FP16 depthwise conv2d & $10^{-3}$ & $10^{-3}$ & 0.95 \\
KleidiAI requantized INT8 GEMM (SME2) & $0.5$ & $10^{-2}$ & 0.99 \\
KleidiAI UINT8 GEMM, FP32 output (SME2) & $10^{-3}$ & $10^{-5}$ & 1.00 \\
\midrule
 & \multicolumn{3}{c}{Minimum SQNR} \\
\cmidrule(l){2-4}
llama.cpp q4\_K\_M GEMM and MoE; q8\_0 MoE & \multicolumn{3}{c}{20 dB} \\
KleidiAI INT4 GEMM (SVE) & \multicolumn{3}{c}{30 dB} \\
KleidiAI INT4 GEMM (SME2) & \multicolumn{3}{c}{40 dB} \\
KleidiAI BF16 GEMM (SME2) & \multicolumn{3}{c}{45 dB} \\
Other KleidiAI INT8, INT4, and INT2 GEMMs & \multicolumn{3}{c}{expert's SQNR on the workload $-$ 1 dB} \\
\bottomrule
\end{tabular}
\end{table}

\subsection{Human effort in one extension}
\label{app:extension-cost}
To show what extending the benchmark asks of humans, we logged one extension of the end-to-end
task set (Appendix~\ref{app:e2e}): rebuilding the workloads of the Qwen3.5-4B and
Llama-3.1-8B-Instruct GEMM tasks from model runs, and collecting workloads for two OLMoE GEMM
tasks. The construction agent, Claude Code with Claude Sonnet 5.5, ran every step through \texttt{kgen}, the command-line tool that
bundles the scripts of Appendix~\ref{app:skills} in the released code. Table~\ref{tab:extension-cost}
lists what the agent and the humans did at each step. The timed commands took about 23 minutes,
and the Graviton instances were held for about 4.5 hours, including setup, model downloads, and
reruns.

\begin{table}[!ht]
\revall
\caption{\revall \textbf{Division of work in one extension} of the end-to-end GEMM tasks.}
\label{tab:extension-cost}
\centering\footnotesize
\begin{tabular}{@{}>{\raggedright\arraybackslash}p{0.2\linewidth}>{\raggedright\arraybackslash}p{0.39\linewidth}>{\raggedright\arraybackslash}p{0.34\linewidth}@{}}
\toprule
Step & Agent & Human \\
\midrule
Scope and definitions & Writes one definition per shape and quantization format & Chooses the models and formats; corrects two shapes the agent got wrong \\
Workloads (\texttt{collect}, \texttt{add}) & Runs each model, records the shapes every GEMM sees, selects and registers workloads & --- \\
Inputs (\texttt{capture}) & Stores the activations reaching each GEMM as workload inputs & --- \\
Calibration (\texttt{calibrate}) & Measures the SQNR of ggml's own arithmetic on every workload & --- \\
Machines & Provisions instances, sets them up, reruns failed steps & Chooses the instance type and approves each launch \\
\bottomrule
\end{tabular}
\end{table}

The human sent 37 messages during the session. 17 changed the course of the work: 14
decisions and 3 corrections. The rest were questions. The decisions fall into two kinds:
choosing what to build, such as covering both quantization formats and splitting definitions by
format, and choosing/approving machines. 

\section{Evaluation Infrastructure}
\label{app:infrastructure}
\subsection{Target machines}
For each run, the control plane provisions a dedicated target machine (an AWS Graviton instance, or a
\texttt{mac-m4.metal} host for M4), evaluates the tasks sequentially with retries on failure, and
collects each task's trajectory. Table~\ref{tab:generations} lists the environment of each hardware generation. No hardware documentation is given to the agent in the main runs, and the
documentation ablation adds the Arm Neoverse V1/V2 optimization guides and an SVE intrinsics
summary on SVE, and SME programming notes on SME2.
\label{app:generations}
\input{tables/generations}
On M4, performance and efficiency cores cannot be selected separately, so comparisons
between Neon and SME2 use the same M4 host. Cross-generation comparisons use only shared
kernels.

\subsection{Instruction-set ladders}
\label{app:isa-ladder}
Table~\ref{tab:isa-arms} gives the configuration of each arm of the instruction-set ladder
experiment (Table~\ref{tab:isa-ladder}). 
On Graviton4, an arm is set by the compile target.
On Apple M4, no compiler flag separates streaming SVE from ZA, because
enabling SME also enables ZA. The Neon-only and streaming-SVE arms are therefore enforced by
rejecting disallowed constructs in the submitted source and disallowed instructions in the
disassembly of the compiled kernel. 
Within a machine, every arm is scored against the same expert reference, so differences between arms reflect only the
instructions available to the agent. The task set for each ladder is listed in
Table~\ref{tab:task-sets}.

\begin{table}[!ht]
\caption{\textbf{Configuration of the instruction-set ladder arms} (Table~\ref{tab:isa-ladder}).}
\label{tab:isa-arms}
\centering\footnotesize
\setlength{\tabcolsep}{4pt}
\begin{tabular}{@{}>{\raggedright\arraybackslash}p{2.25cm}l>{\raggedright\arraybackslash}p{3.95cm}>{\raggedright\arraybackslash}p{5.1cm}@{}}
\toprule
Arm & Machine & Compile target & Additionally rejected \\
\midrule
L1: Neon only & Graviton4 & \texttt{-march=armv8-a} & --- (SVE intrinsics do not compile) \\
L2: $+$ SVE & Graviton4 & \texttt{-march=armv8.2-a+sve} & --- (SVE2 intrinsics do not compile) \\
L3: $+$ SVE2 & Graviton4 & \texttt{-march=armv9-a+sve2} & --- \\
\cmidrule{1-4}
L4: Neon only & Apple M4 & \texttt{-mcpu=apple-m4+nosme} & Source: SVE and SME headers and types, streaming attributes, \texttt{smstart}. Binary: \texttt{smstart}, SVE and ZA/ZT0 registers. \\
L5: $+$ streaming SVE & Apple M4 & \texttt{-mcpu=apple-m4} & Source: ZA/ZT0 attributes and intrinsics, \texttt{smstart za}. Binary: \texttt{smstart} without \texttt{sm}, outer products (\texttt{MOPA}/\texttt{MOPS}), ZA/ZT0 registers. \\
L6: $+$ ZA (full SME2) & Apple M4 & \texttt{-mcpu=apple-m4} & --- \\
\bottomrule
\end{tabular}
\end{table}

\subsection{Agent interface, harnesses, and prompts}
\label{app:interface}
\label{app:harness}
\label{app:prompts}
\paragraph{What the agent sees.} For each task, the agent receives the kernel specification in
code form: the definition name and operator type, including the constant attributes (e.g., kernel size and stride, or $N$ and $K$ of a
GEMM), and the scalar starting implementation. It does not receive the
workloads: the values of the variable dimensions (e.g., channel counts, spatial sizes, or token
counts), the number of workloads, and the input tensors are never shown. The kernel receives the
variable dimensions as run-time arguments, so a candidate must be correct for every size that the
specification allows. Feedback is aggregated over the hidden workloads: \emph{evaluate} reports
the maximum errors, the geometric-mean speedup, and hardware counters where available,
and after a failure only the identifier of the first failing workload and an error message. The
agent also sees the results of its own earlier calls. Every \emph{compile}, \emph{evaluate}, and
\emph{disassemble} call counts toward the 40-call budget. A run counts as complete only after at
least 15 calls, and the best correct candidate is retained.


All harnesses receive the same task skill. Claude Code and Codex run in a
fresh, empty working directory with shell and web tools disabled and auto-memory off. Nanobot uses temperature 0.1, reasoning effort \texttt{xhigh}, and a
200k-token context, with shell and web tools disabled and file access limited to the task's
workspace. The tool-call loop uses temperature 0.2 if available. In
long runs, the tool-call loop replaces earlier kernel versions with a summary of the versions that
passed, and nanobot compacts older tool results when its context fills up.

Figure~\ref{fig:agent-prompt} shows the prompt given to the agent and
Table~\ref{tab:agent-tools} the tools it can call.

\begin{figure}[!ht]
\begin{AIbox}{Agent Prompt}
\small
{\color{blue}\bf System (kernel-session skill, condensed):}

\textbf{Ground Rules}

\verb|  |* Work on one definition at a time; \emph{evaluate} and \emph{disassemble} must name the definition and version of your last \emph{compile}.

\verb|  |* Stay on the target ISA; do not fall back to an older one.

\verb|  |* Never use OpenMP or other parallelization. Kernels always run on the remote target, which has every dependency installed.

\verb|  |* Use \emph{disassemble} to check whether SIMD is really helping and to find bottlenecks.

\textbf{Workflow}

\verb|  |0. Call \emph{check\_progress} to look up the best result so far. If one exists, compile and evaluate it first, then continue from it.

\verb|  |1. Read the \texttt{reference-scalar-kernel.cpp} resource, compile it unchanged (this becomes v1), and evaluate it to record the starting point.

\verb|  |2. Iterate: compile an attempt; evaluate it (correctness and timing in one call; the best version is saved automatically); disassemble when IPC is low or the speedup is unexpectedly poor. Re-read earlier versions and the trajectory instead of re-evaluating them.

\textbf{Metrics from evaluate}

\verb|  |* \texttt{max\_absolute\_error}, \texttt{max\_relative\_error}; \texttt{time\_speedup\_geomean} and \texttt{cycle\_speedup\_geomean} against the expert baseline (geometric mean over workloads); \texttt{ipc\_mean}; \texttt{cache\_misses\_mean}.

\verb|  |* On failure, \texttt{failed\_workload} and \texttt{log} say which workload failed and why.

{\color{blue}\bf User (task prompt):}

Optimize the ``\texttt{<definition>}'' kernel definition (dataset: \texttt{<source>}, baseline solution source: \texttt{<expert>}) in ISA \texttt{<isa>}. You must spend at least 15 tool calls but not exceed 40 tool calls to explore genuinely different optimization attempts before you are allowed to submit. Once you hit that ceiling, stop iterating and submit your best version immediately, since every iteration spends real model API budget. Follow the ground rules and workflow in your system prompt.
\end{AIbox}
\caption{\textbf{Agent prompt.} The skill (about 1{,}100 words) is condensed; the task prompt is
verbatim.}
\label{fig:agent-prompt}
\end{figure}

\begin{table}[!ht]
\caption{\textbf{Agent tools}, condensed from their schemas. \emph{definition} and \emph{version}
must match the last \emph{compile}.}
\label{tab:agent-tools}
\centering\footnotesize
\setlength{\tabcolsep}{3pt}
\begin{tabular}{@{}l>{\raggedright\arraybackslash}p{2.3cm}>{\raggedright\arraybackslash}p{5.1cm}>{\raggedright\arraybackslash}p{3.2cm}@{}}
\toprule
Tool & Arguments & What it does & Returns \\
\midrule
\texttt{check\_progress} & \texttt{definition} & Looks up the best result so far on this definition. & Its version, metrics, and code. \\
\addlinespace
\texttt{compile} & \texttt{definition}, \texttt{code} & Compiles the agent's \texttt{kernel.cpp}; the harness and binding files are added automatically. & A version number, or the compiler error. \\
\addlinespace
\texttt{evaluate} & \texttt{definition}, \texttt{version} & Checks correctness on all workloads, stopping at the first failure; if correct, times the kernel. A new best result is saved immediately. & Errors, speedups, and counters; or the failing workload's ID and a log. \\
\addlinespace
\texttt{disassemble} & \texttt{definition}, \texttt{version}, optional \texttt{fn} & Disassembles the kernel entry symbol, or \texttt{fn}. & Up to 300 lines of AArch64 assembly. \\
\bottomrule
\end{tabular}
\end{table}

\subsection{Measurement protocol and integrity}
\label{app:measurement}
Each workload is first run for correctness and then timed in 50 samples after 10 warmup samples, and
the primary metric is the minimum sample. On every target, each sample repeats the kernel until it
lasts at least 10\,ms, timed with a nanosecond-resolution clock, and a candidate uses the same
repetition count and settings as the baseline. On Linux,
the timed process is pinned to one CPU. When available, hardware counters are also collected. Expert baselines are measured on
the same machine before agent evaluation. Per-workload speedup is expert-reference time divided
by candidate time, and task speedup is the geometric mean over workloads. Each \emph{evaluate}
call runs in a fresh subprocess and is stopped if checking and timing all workloads takes longer
than 1200\,s. Re-measuring selected kernels on new instances changes a kernel's speedup by a median of
about $1\%$ and does not change the ranking beyond ties.

Across three seeds of nanobot with GPT-5.6-luna
on Graviton4, on the 32-kernel subset,
Speedup@40 varies by 4.5\%, and per-kernel Speedup@40 by a median of 9.7\%. Re-timing the same
fixed code varies by only 0.2\%, so run-to-run variation comes from agent stochasticity rather
than from measurement. \rev{Over three seeds, a column's geometric mean in Table~\ref{tab:main-sve} varies by $3$--$5\%$ (one s.d.), so rows within this margin are tied.} %

\rev{Table~\ref{tab:ci} gives bootstrap $95\%$ confidence intervals over kernels for every cell of
Table~\ref{tab:main-sve}. They agree with the seed variation: in paired comparisons with the best
system of each column, $2$, $5$, $8$, and $0$ of the other $13$ systems are not separable on ncnn,
llama.cpp, SIMD Loops, and KleidiAI.}
\begin{table}[!ht]
\revall
\caption{\revall \textbf{Bootstrap 95\% confidence intervals} (10{,}000 resamples of the kernels) of the
geometric means in Table~\ref{tab:main-sve}. KleidiAI has six kernels, so its intervals are wide.}
\label{tab:ci}
\centering\small
\resizebox{\linewidth}{!}{%
\begin{tabular}{llcccc}
\toprule
Harness & Model & ncnn (23) & llama.cpp (26) & SIMD Loops (24) & KleidiAI (6) \\
\midrule
Claude Code    & Claude Opus 4.8   & 2.574\,{\footnotesize[1.96, 3.40]} & 1.779\,{\footnotesize[1.26, 2.61]} & 1.555\,{\footnotesize[1.24, 2.09]} & 0.701\,{\footnotesize[0.45, 0.98]} \\
Claude Code    & Claude Sonnet 4.6 & 1.934\,{\footnotesize[1.50, 2.55]} & 1.243\,{\footnotesize[0.88, 1.82]} & 1.516\,{\footnotesize[1.20, 2.06]} & 0.386\,{\footnotesize[0.20, 0.75]} \\
Codex          & GPT-5.6-sol       & 1.815\,{\footnotesize[1.37, 2.39]} & 1.297\,{\footnotesize[0.94, 1.82]} & 1.487\,{\footnotesize[1.19, 1.96]} & 0.431\,{\footnotesize[0.23, 0.74]} \\
Codex          & GPT-5.6-luna      & 0.674\,{\footnotesize[0.47, 0.97]} & 0.627\,{\footnotesize[0.38, 1.02]} & 1.127\,{\footnotesize[0.89, 1.45]} & 0.147\,{\footnotesize[0.06, 0.36]} \\
nanobot        & Claude Opus 4.8   & 2.597\,{\footnotesize[2.01, 3.38]} & 1.708\,{\footnotesize[1.22, 2.48]} & 1.521\,{\footnotesize[1.21, 2.02]} & 0.453\,{\footnotesize[0.24, 0.83]} \\
nanobot        & Claude Sonnet 4.6 & 2.046\,{\footnotesize[1.53, 2.78]} & 1.449\,{\footnotesize[1.01, 2.12]} & 1.503\,{\footnotesize[1.21, 1.97]} & 0.231\,{\footnotesize[0.12, 0.48]} \\
nanobot        & GPT-5.6-sol       & 2.156\,{\footnotesize[1.62, 2.89]} & 1.334\,{\footnotesize[0.96, 1.91]} & 1.543\,{\footnotesize[1.22, 2.09]} & 0.319\,{\footnotesize[0.17, 0.59]} \\
nanobot        & GPT-5.6-luna      & 1.110\,{\footnotesize[0.79, 1.55]} & 0.878\,{\footnotesize[0.59, 1.36]} & 1.182\,{\footnotesize[0.93, 1.49]} & 0.161\,{\footnotesize[0.07, 0.39]} \\
nanobot        & DeepSeek-V4-Pro   & 1.091\,{\footnotesize[0.77, 1.55]} & 0.850\,{\footnotesize[0.53, 1.36]} & 1.458\,{\footnotesize[1.15, 1.97]} & 0.231\,{\footnotesize[0.09, 0.59]} \\
nanobot        & GLM-5.3           & 2.743\,{\footnotesize[2.05, 3.68]} & 1.954\,{\footnotesize[1.38, 2.85]} & 1.545\,{\footnotesize[1.21, 2.14]} & 0.499\,{\footnotesize[0.25, 0.85]} \\
tool-call loop & Claude Opus 4.8   & 2.308\,{\footnotesize[1.71, 3.18]} & 1.828\,{\footnotesize[1.34, 2.62]} & 1.480\,{\footnotesize[1.22, 1.86]} & 0.496\,{\footnotesize[0.27, 0.86]} \\
tool-call loop & Claude Sonnet 4.6 & 2.043\,{\footnotesize[1.54, 2.78]} & 1.915\,{\footnotesize[1.41, 2.66]} & 1.521\,{\footnotesize[1.20, 2.09]} & 0.297\,{\footnotesize[0.12, 0.72]} \\
tool-call loop & GPT-5.6-sol       & 1.999\,{\footnotesize[1.51, 2.72]} & 1.671\,{\footnotesize[1.24, 2.32]} & 1.513\,{\footnotesize[1.19, 2.09]} & 0.307\,{\footnotesize[0.15, 0.65]} \\
tool-call loop & GPT-5.6-luna      & 0.955\,{\footnotesize[0.68, 1.35]} & 0.854\,{\footnotesize[0.57, 1.31]} & 1.245\,{\footnotesize[1.02, 1.55]} & 0.164\,{\footnotesize[0.06, 0.47]} \\
\bottomrule
\end{tabular}}
\end{table}

\section{\texorpdfstring{\rev{Speedup over the Starting Point and Fraction of Hardware Peak}}{Speedup over the Starting Point and Fraction of Hardware Peak}}
\label{app:baselines}
\rev{Every agent starts from the scalar reference compiled with \texttt{-O3} and auto-vectorization for
SVE (the ``scalar + \texttt{-O3}, SVE enabled'' row of Table~\ref{tab:main-sve}).
Table~\ref{tab:vs-start} reports every cell of Table~\ref{tab:main-sve} against this starting point
instead of the expert. Every system is faster than its starting point in every column}
($2.4$--$19.8\times$), while the experts are $2.1$--$28.2\times$ faster.

\begin{table}[!ht]
\revall
\caption{\revall \textbf{Speedup over the starting point} (scalar reference, \texttt{-O3}, SVE enabled) on
Graviton3, geometric mean over kernels and, in parentheses, the fraction of kernels faster than
the starting point. Same runs and task sets as Table~\ref{tab:main-sve}.}
\label{tab:vs-start}
\centering\small
\resizebox{\linewidth}{!}{%
\begin{tabular}{llcccc}
\toprule
Harness & Model & ncnn (23) & llama.cpp (26) & SIMD Loops (24) & KleidiAI (6) \\
\midrule
Claude Code    & Claude Opus 4.8   & 11.24\,{\footnotesize(100\%)} & 14.18\,{\footnotesize(100\%)} & 3.32\,{\footnotesize(92\%)} & 19.77\,{\footnotesize(100\%)} \\
Claude Code    & Claude Sonnet 4.6 & 8.44\,{\footnotesize(100\%)} & 9.91\,{\footnotesize(100\%)} & 3.24\,{\footnotesize(92\%)} & 10.88\,{\footnotesize(100\%)} \\
Codex          & GPT-5.6-sol       & 7.92\,{\footnotesize(100\%)} & 10.34\,{\footnotesize(100\%)} & 3.18\,{\footnotesize(88\%)} & 12.14\,{\footnotesize(100\%)} \\
Codex          & GPT-5.6-luna      & 2.94\,{\footnotesize(100\%)} & 5.00\,{\footnotesize(100\%)} & 2.41\,{\footnotesize(79\%)} & 4.15\,{\footnotesize(100\%)} \\
nanobot        & Claude Opus 4.8   & 11.34\,{\footnotesize(100\%)} & 13.62\,{\footnotesize(100\%)} & 3.25\,{\footnotesize(88\%)} & 12.77\,{\footnotesize(100\%)} \\
nanobot        & Claude Sonnet 4.6 & 8.93\,{\footnotesize(100\%)} & 11.55\,{\footnotesize(100\%)} & 3.21\,{\footnotesize(92\%)} & 6.52\,{\footnotesize(100\%)} \\
nanobot        & GPT-5.6-sol       & 9.41\,{\footnotesize(100\%)} & 10.63\,{\footnotesize(100\%)} & 3.30\,{\footnotesize(92\%)} & 8.98\,{\footnotesize(100\%)} \\
nanobot        & GPT-5.6-luna      & 4.85\,{\footnotesize(100\%)} & 7.00\,{\footnotesize(100\%)} & 2.53\,{\footnotesize(88\%)} & 4.54\,{\footnotesize(100\%)} \\
nanobot        & DeepSeek-V4-Pro   & 4.76\,{\footnotesize(100\%)} & 6.77\,{\footnotesize(100\%)} & 3.12\,{\footnotesize(92\%)} & 6.52\,{\footnotesize(100\%)} \\
nanobot        & GLM-5.3           & 11.97\,{\footnotesize(100\%)} & 15.57\,{\footnotesize(96\%)} & 3.30\,{\footnotesize(92\%)} & 14.06\,{\footnotesize(100\%)} \\
tool-call loop & Claude Opus 4.8   & 10.07\,{\footnotesize(100\%)} & 14.57\,{\footnotesize(100\%)} & 3.16\,{\footnotesize(92\%)} & 13.98\,{\footnotesize(100\%)} \\
tool-call loop & Claude Sonnet 4.6 & 8.92\,{\footnotesize(100\%)} & 15.26\,{\footnotesize(100\%)} & 3.25\,{\footnotesize(92\%)} & 8.38\,{\footnotesize(100\%)} \\
tool-call loop & GPT-5.6-sol       & 8.73\,{\footnotesize(100\%)} & 13.32\,{\footnotesize(100\%)} & 3.23\,{\footnotesize(88\%)} & 8.67\,{\footnotesize(100\%)} \\
tool-call loop & GPT-5.6-luna      & 4.17\,{\footnotesize(100\%)} & 6.80\,{\footnotesize(100\%)} & 2.66\,{\footnotesize(88\%)} & 4.62\,{\footnotesize(100\%)} \\
\midrule
\multicolumn{2}{l}{Expert reference} & 4.37 & 7.97 & 2.14 & 28.20 \\
\bottomrule
\end{tabular}}
\end{table}

Table~\ref{tab:peak} gives the fraction of the theoretical limit, $t_{\mathrm{floor}} /
t_{\mathrm{measured}}$ per workload with $t_{\mathrm{floor}} = \max(W/P,\, Q/B)$: $W$ operations at the
measured peak $P$ of the fastest instruction for the data type, or $Q$ bytes at the bandwidth $B$ of
the cache level that holds the data. The ncnn and llama.cpp experts reach $7$--$8\%$ of the limit
and the frontier agents two to three times that, while the KleidiAI experts ($43\%$) stay ahead of
every agent. \textbf{Therefore, a speedup above 1 shows headroom in the reference, not that the agent is near
the hardware limit.}

\begin{table}[!ht]
\revall
\caption{\revall \textbf{Fraction of the theoretical hardware limit} reached on Graviton3 (single core),
geometric mean over workloads and kernels; only the expert is timed, and the other rows scale by
their speedups. SIMD Loops: the 21 kernels with a defined operation count. KleidiAI: the six
kernels, with the expert and the cache bandwidths measured again on Graviton3.}
\label{tab:peak}
\centering\small
\resizebox{\linewidth}{!}{%
\begin{tabular}{llcccc}
\toprule
Harness & Model & ncnn (23) & llama.cpp (26) & SIMD Loops (21) & KleidiAI (6) \\
\midrule
Claude Code    & Claude Opus 4.8   & 17.4\% & 14.2\% & 50.0\% & 30.1\% \\
Claude Code    & Claude Sonnet 4.6 & 13.1\% & 9.9\% & 48.9\% & 16.5\% \\
Codex          & GPT-5.6-sol       & 12.3\% & 10.4\% & 48.2\% & 18.5\% \\
Codex          & GPT-5.6-luna      & 4.6\% & 5.0\% & 38.4\% & 6.3\% \\
nanobot        & Claude Opus 4.8   & 17.6\% & 13.7\% & 49.0\% & 19.4\% \\
nanobot        & Claude Sonnet 4.6 & 13.9\% & 11.6\% & 48.6\% & 9.9\% \\
nanobot        & GPT-5.6-sol       & 14.6\% & 10.7\% & 49.3\% & 13.7\% \\
nanobot        & GPT-5.6-luna      & 7.5\% & 7.0\% & 39.7\% & 6.9\% \\
nanobot        & DeepSeek-V4-Pro   & 7.4\% & 6.8\% & 46.3\% & 9.9\% \\
nanobot        & GLM-5.3           & 18.6\% & 15.6\% & 49.0\% & 21.4\% \\
tool-call loop & Claude Opus 4.8   & 15.6\% & 14.7\% & 49.4\% & 21.3\% \\
tool-call loop & Claude Sonnet 4.6 & 13.8\% & 15.3\% & 48.2\% & 12.7\% \\
tool-call loop & GPT-5.6-sol       & 13.5\% & 13.3\% & 47.7\% & 13.2\% \\
tool-call loop & GPT-5.6-luna      & 6.5\% & 6.8\% & 41.5\% & 7.0\% \\
\midrule
\multicolumn{2}{l}{Starting point (\texttt{-O3}, SVE)} & 1.6\% & 1.0\% & 17.4\% & 1.5\% \\
\multicolumn{2}{l}{Expert reference} & 6.8\% & 8.0\% & 37.4\% & 42.9\% \\
\bottomrule
\end{tabular}}
\end{table}

\section{Case Studies}
\label{app:casestudies}
\subsection{End-to-end optimization with agent-written kernels}
\label{app:e2e}

\paragraph{Setup.}
The main experiments measure adaptation one kernel at a time. To test whether
these gains transfer to an application, we test two popular open-weight
models, Qwen3.5-4B and Llama-3.1-8B-Instruct (both Q4\_K\_M), served with llama.cpp
on Graviton4. 11 individual quantized matrix-multiplication kernels from the Qwen model
and 7 matrix-multiplication kernels from the Llama model are turned
into benchmark tasks. We substitute the resulting agent-written kernels into
the inference stack and compare against original llama.cpp on the same
hardware.

\paragraph{Model selection.} Qwen3.5-4B is a recent hybrid model (Gated DeltaNet and attention);
Llama-3.1-8B-Instruct is a standard dense transformer widely used for CPU inference. Both use
4-bit quantization, the standard for CPU inference in llama.cpp.
  \begin{table}[!ht]
  \revall
  \centering
  \begin{minipage}[t]{0.48\linewidth}
  \centering
  \caption{\textbf{End-to-end speedup and perplexity for Qwen3.5-4B}
  (Claude Code + Claude          
  Fable 5.1), relative to original llama.cpp.}
  \label{tab:e2e-details}
  \footnotesize
  \begin{tabular*}{\linewidth}{@{\extracolsep{\fill}}lcc}
  \toprule
  Threads & Prefill & Decode \\
  \midrule
   \multicolumn{3}{@{}l}{\emph{Agent kernels}} \\

  1  & 1.395 & 1.045 \\
  4  & 1.348 & 1.059 \\
  16 & 1.298 & 1.048 \\
  \multicolumn{3}{@{}l}{\emph{Original without repack}} \\
  1  & 0.460 & 0.758 \\
  4  & 0.465 & 0.744 \\
  16 & 0.521 & 0.879 \\
  \midrule
  Model & Perplexity & $\Delta$ \\
  \midrule
  BF16             & 9.384 & $-$1.3\% \\
  Q6\_K            & 9.436 & $-$0.7\% \\
  Q4\_K\_M         & 9.506 & --- \\
  Q4\_K\_M (agent kernels) & \textbf{9.504} & 0.0\% \\
  Q3\_K\_M         & 9.846 & $+$3.6\% \\
  \bottomrule
  \end{tabular*}
  \end{minipage}\hfill
  \begin{minipage}[t]{0.48\linewidth}
  \centering
  \caption{\textbf{Prefill speedup by prompt length and micro-batch} at
  16 threads,
  relative to original llama.cpp.}
  \label{tab:e2e-sweeps}
  \footnotesize
  \begin{tabular*}{\linewidth}{@{\extracolsep{\fill}}lcc}
  \toprule
  Setting & Qwen3.5-4B & Llama-3.1-8B \\
  \midrule
   \multicolumn{3}{@{}l}{\emph{Prompt length}} \\
  512  & 1.298 & 1.393 \\
  2048 & 1.281 & 1.322 \\
  4096 & 1.260 & 1.255 \\
  8192 & 1.226 & 1.184 \\
  \midrule
  \multicolumn{3}{@{}l}{\emph{Micro-batch size}} \\
  1   & 1.053 & 1.090 \\
  2   & 1.270 & 1.598 \\
  8   & 1.164 & 1.479 \\
  64  & 1.061 & 0.989 \\
  512 & 1.298 & 1.397 \\
  \bottomrule
  \end{tabular*}
  \end{minipage}
  \end{table}
\paragraph{End-to-end performance and quality.}
  Table~\ref{tab:e2e-details} reports the speedup across different thread
  counts and quantizations relative to llama.cpp for Qwen3.5-4B, and
  Table~\ref{tab:e2e-sweeps} reports speedup across prompt length and
  micro-batch size for both models. For Qwen3.5-4B, the
  adapted kernels speed up prefill by $1.30\times$ at 16 threads and by
  $1.35$--$1.39\times$ at 1--4 threads. Decode improves by
  $1.04$--$1.06\times$, consistent with its greater sensitivity to memory
  bandwidth and
  to operators that are not replaced. Perplexity is unchanged relative
  to the Q4\_K\_M model (9.504 vs.\ 9.506). For Llama-3.1-8B-Instruct, the agent
  kernels speed up prefill by
  $1.39$--$1.57\times$ and decode by $1.07$--$1.10\times$ across 1--16
  threads
  (Table~\ref{tab:e2e}), and perplexity is unchanged (7.879 in both
  builds).

  \paragraph{Serving regimes.}
  The prefill gain persists across prompt lengths and micro-batch sizes
  (Table~\ref{tab:e2e-sweeps}). It declines with prompt length as the share of work in unreplaced
  operators, mainly attention, grows, and batch size 1 behaves like decode.

\paragraph{Adaptation to the hardware.}
All 18 kernels use Graviton4's i8mm
matrix-multiply instructions. The agent
enables these instructions per function and combines them with shape-specific
tiling, layout, and prefetching. Measured alone, agent kernels are $1.45\times$ faster for Qwen and $1.62\times$ for Llama than llama.cpp's repacked kernels. Each generated kernel serves one fixed shape,
whereas a general-purpose library implementation must cover a broader shape
distribution. We note that this is not a general claim that agents out-engineer expert libraries end-to-end, but rather demonstrates the potential of
hardware-specific kernel generation.

\subsection{Learning SME2 at test time}
\label{app:sme2-case}
Table~\ref{tab:sme2-traj} lists every counted tool call of one run: nanobot with Claude Opus 4.8
on KleidiAI's BF16 GEMM on Apple M4, the run behind the $0.94\times$ result in
Section~\ref{sec:adapt}.

The agent starts from the scalar implementation ($0.004\times$ the KleidiAI SME2 kernel) and
targets the matrix engine directly, planning a \texttt{BFMOPA} outer-product kernel that
accumulates into the ZA array. Its first attempt does not compile because it gathers operands
inside streaming mode, where gather loads are unavailable, so the agent packs both operands to
BF16 in non-streaming code. The next two kernels run but are incorrect, until the agent works out
that the governing predicates of \texttt{BFMOPA} are defined over the 16-bit source elements
rather than the 32-bit outputs. Six calls thus pass before the first correct SME2 kernel
($0.37\times$). Blocking the output over all four FP32 ZA tiles then more than doubles the
speedup ($0.80\times$). The remaining calls are ordinary performance engineering: the agent finds
that packing the operands, not the outer products, limits performance, and cache-blocked Neon
packing brings the kernel to $0.935\times$ at call 29.

\input{tables/sme2-traj}

\section{Use of Target-Specific Intrinsics}
\label{app:intrinsic-usage}
We check whether agents actually use the instructions of the target extension. For each run, the
final solution is the version whose result is reported as Speedup@40. A final
\rev{solution counts as using SVE if it calls at least one SVE intrinsic, and as using SME2 if its
disassembled binary contains at least one ZA instruction, which also covers instructions used
through inline assembly. Every solution we count as using an extension also has that extension's
instructions in its binary.}
\rev{Table~\ref{tab:intrinsic-usage} reports the
nanobot runs of the four models}: the SVE kernels of Table~\ref{tab:main-sve}, the SVE2 loops of
Table~\ref{tab:evolve}, and the six KleidiAI SME2 kernels of the same table in each of its three seeds.
Nearly all final SVE solutions use SVE
instructions, but on SME2 only Claude Opus 4.8 uses the matrix tiles in most of its final solutions. %
This is consistent with the drop from SVE to SME2 in Table~\ref{tab:evolve}.
On the eight SIMD Loops kernels of the SVE2 comparison, a solution counts as using SVE2 if it does
not compile without SVE2, the test used to select these loops. Most Claude solutions do, and in 13
of the 14 cases the solution uses the same SVE2 instruction as the expert. 

\begin{table}[!ht]
\revall
\caption{\revall \textbf{Final solutions that use intrinsics of the target extension} (nanobot): SVE
intrinsics on SVE (79 kernels), SVE2-only instructions on SVE2 (8 SIMD Loops kernels), ZA
instructions in the binary on SME2 (6 KleidiAI kernels, 3 seeds).}
\label{tab:intrinsic-usage}
\centering\footnotesize
\begin{tabular}{lcccc}
\toprule
Target & Claude Opus 4.8 & Claude Sonnet 4.6 & DeepSeek-V4-Pro & GLM-5.3 \\
\midrule
SVE  & 79/79 (100\%) & 75/79 (95\%) & 71/79 (90\%) & 78/79 (99\%) \\
SVE2 & 6/8 (75\%) & 8/8 (100\%) & 4/8 (50\%) & 3/8 (38\%) \\
SME2 & 15/18 (83\%) & 5/18 (28\%) & 1/18 (6\%) & 5/18 (28\%) \\
\bottomrule
\end{tabular}
\end{table}

\section{\texorpdfstring{\rev{SIMD class of the SME2 kernels on Apple M4}}{SIMD class of the SME2 kernels on Apple M4}}
\label{app:m4-ladder-troubleshooting}
This section examines which SIMD class the agent uses on the 20 KleidiAI SME2 kernels on Apple M4.
A version is ZA if it uses ZA tile intrinsics (\texttt{svmopa}, \texttt{\_za} intrinsics or \texttt{\_\_arm\_new}), streaming SVE if it uses \texttt{\_\_arm\_streaming} or \texttt{\_\_arm\_locally\_streaming} without ZA, Neon if it uses Neon intrinsics outside streaming mode, and plain C otherwise. The final kernel of a run is its best passing version within the first 40 tool calls.

\hide{
\subsection{\rev{Best-fit class per kernel}}
\rev{A kernel's \emph{best fit} is the class of its fastest passing version across four runs. The best fit is ZA for 9 kernels and Neon for 11, and never streaming SVE alone. On 15 of the 20 kernels, the default SME2 run ends on a different class (Table~\ref{tab:sme2-simd-fit}). The agent never ends on a working ZA version: it attempts ZA on only 2 of the 9 ZA kernels, both attempts fail to build, and these kernels stay up to 23$\times$ slower than their best fit. Of the 11 Neon kernels, 2 stay in streaming SVE without trying Neon, and 4 end on plain C, either because streaming SVE never beats the scalar starting point (2) or because every SIMD version fails correctness (2, fp16 and int4 decoding). The agent also rarely revisits its first choice: on 11 of the 20 kernels it tries only one class (10 streaming SVE, 1 Neon).}

\begin{table}[h]
\revall
\centering
\small
\caption{\revall Best-fit SIMD class per kernel vs.\ the class the default SME2 run ends on.}
\label{tab:sme2-simd-fit}
\begin{tabular}{lrlr}
\toprule
Best fit & Kernels & Agent ended on & Mismatches \\
\midrule
ZA                     & 9  & streaming SVE 4, plain C 3, Neon 2 & 9 \\
Neon                   & 11 & Neon 5, plain C 4, streaming SVE 2 & 6 \\
Streaming SVE (no ZA)  & 0  & --                                 & -- \\
\midrule
Total                  & 20 &                                    & 15 \\
\bottomrule
\end{tabular}
\end{table}
}

\label{app:ladder-simd-class}
\label{app:ladder-compile}
Table~\ref{tab:ladder-simd-class} classifies the final kernel of every (kernel, seed) run on the ladder of Table~\ref{tab:isa-ladder}, normalized by the same kernel on the Neon-only tier (its speedup over the expert divided by its mean over the three Neon-only seeds). On the streaming-SVE tier, the agent keeps Neon in only 1 of 60 runs, and its streaming-SVE and plain-C kernels reach only 0.37 and 0.42 of Neon-only. For the full-SME2 tier, across its 60 runs the agent writes ZA in only two versions, and neither passes. Compile errors also grow with the exposed ISA: on the first seed, 48 of 374 compiles fail (13\%) on the Neon-only tier, 82 of 405 (20\%) on the streaming-SVE tier, and 102 of 419 (24\%) with full SME2.

\begin{table}[h]
\centering
\small
\begin{tabular}{llrr}
\toprule
Tier & SIMD class of final kernel & \# final kernels (of 60) & Relative to Neon only \\
\midrule
Neon only        & Neon           & 59 & 0.99 \\
                 & plain C        & 1  & 0.16 \\
\midrule
Streaming SVE    & streaming SVE  & 25 & 0.37 \\
                 & plain C        & 34 & 0.42 \\
                 & Neon           & 1  & 0.76 \\
\midrule
Full SME2        & Neon           & 23 & 0.75 \\
                 & streaming SVE  & 17 & 0.40 \\
                 & plain C        & 20 & 0.59 \\
                 & ZA             & 0  & --   \\
\bottomrule
\end{tabular}
\caption{Final kernels on the M4 ladder by SIMD class (20 kernels $\times$ 3 seeds per tier). Last column: geometric mean of the normalized speedup.}
\label{tab:ladder-simd-class}
\end{table}

\section{Contamination Statement}
\label{app:contamination}
Because ncnn, llama.cpp, SIMD Loops, and
  KleidiAI are public, expert code may exist in models' training data.
  However, on most ncnn, llama.cpp, and SIMD Loops kernels, agents exceed
  the expert code rather than reproducing it, and use intrinsics not in library code and different algorithm structures. Further, the task setup
  does not match the library: agents start from portable scalar code with
  benchmark-specific interfaces. Finally, KleidiAI's SME2 expert kernels
  are also public, yet agents fall far short of them.
  \rev{As a direct test, on the 20 KleidiAI SME2 kernels of Table~\ref{tab:isa-ladder} (full SME2 arm), the 6 whose expert
  micro-kernels were added to KleidiAI after the knowledge cutoff of GPT-5.6-luna (2026-02-16) score about the same as
  the 14 older ones (geometric-mean Speedup@40 0.088 vs.\ 0.082). The workloads used for the correctness check and for
  timing are also hidden from the agents.}

%% file: tables/generations.tex
\begin{table}[!t]
\caption{\textbf{Evaluation environment per hardware generation.} Counter feedback lists the counters returned by \emph{evaluate}; reported speedups use wall-clock time.}
\label{tab:generations}
\centering\footnotesize
\setlength{\tabcolsep}{3pt}
\begin{tabular}{@{}l>{\raggedright\arraybackslash}p{2.35cm}>{\raggedright\arraybackslash}p{1.7cm}>{\raggedright\arraybackslash}p{1.55cm}l>{\raggedright\arraybackslash}p{0.9cm}>{\raggedright\arraybackslash}p{1.6cm}@{}}
\toprule
Gen. & Machine & Core, vector width & OS / compiler & Target flag & CPU pinning & Counter feedback \\
\midrule
SVE  & Graviton3, AWS \texttt{c7g.xlarge} & Neoverse V1, 256-bit SVE & Ubuntu 22.04, clang-18 & \texttt{-march=armv8.2-a+sve} & CPU 0 & cycles, instructions, cache misses \\
\addlinespace
SVE2 & Graviton4, AWS \texttt{c8g.xlarge} & Neoverse V2, 128-bit SVE2 & Ubuntu 22.04, clang-18 & \texttt{-march=armv9-a+sve2} & CPU 0 & cycles, instructions, cache misses \\
\addlinespace
SME2 & Apple M4, AWS \texttt{mac-m4.metal} & 512-bit, streaming SVE only & macOS, Apple clang & \texttt{-mcpu=apple-m4} & none & none \\
\bottomrule
\end{tabular}
\end{table}

%% file: tables/sme2-traj.tex
\begin{table}[!ht]
\caption{\textbf{Tool calls of one SME2 run} (KleidiAI \texttt{gemm\_bf16\_n512\_k512}, nanobot with
Claude Opus 4.8, Apple M4). Speedups are relative to the KleidiAI SME2 kernel.}
\label{tab:sme2-traj}
\centering\footnotesize
\begin{tabular}{@{}l>{\raggedright\arraybackslash}p{8.6cm}>{\raggedright\arraybackslash}p{3.4cm}@{}}
\toprule
Calls & What the agent does & Result \\
\midrule
1--2   & compile and evaluate the given scalar code & 0.004$\times$ \\
3      & first \texttt{BFMOPA} kernel, operands gathered inside streaming mode & compile error \\
4--7   & pack operands to BF16 outside streaming mode; one ZA tile; fix load predicates & incorrect (twice) \\
8--9   & 16-bit governing predicates for \texttt{BFMOPA} & 0.370$\times$ \\
10--11 & $2{\times}2$ blocking over all four FP32 ZA tiles & 0.799$\times$ \\
12--20 & Neon packing of $B$, disassembly, alternative packing of $A$, $K$ unrolling & 1 compile error; 0.507--0.799$\times$ \\
21--25 & multi-slice ZA reads in the output store & 1 compile error; 0.838--0.839$\times$ \\
26--29 & cache-blocked Neon packing of $A$, wider packing of $B$ (best) & \textbf{0.935$\times$} \\
30--35 & wider loads, branch hoisting, loop reordering & 0.925--0.935$\times$ \\
36     & one more packing variant, compiled but not evaluated before the run ends & --- \\
\bottomrule
\end{tabular}
\end{table}

%% file: references.bib
@misc{openai2026gpt56,
  author = {Matthew Ferrari and Philippe Tillet and Ahmed Ibrahim and Joe Gershenson and Steve Coffey},
  title = {How {GPT-5.6} Fuses Frontier Intelligence with Frontier Efficiency},
  year = {2026},
  month = jul,
  howpublished = {OpenAI, \url{https://openai.com/index/gpt-5-6-frontier-intelligence-efficiency/}}
}

@misc{arm2026acle,
  author = {{Arm}},
  title = {{Arm C Language Extensions}},
  year = {2026},
  howpublished = {\url{https://github.com/ARM-software/acle/releases/tag/r2026Q1}},
  note = {ACLE 2026 Q1}
}

@inproceedings{remke2024hellosme,
  author = {Stefan Remke and Alexander Breuer},
  title = {Hello {SME}! {Generating} Fast Matrix Multiplication Kernels Using the {Scalable Matrix Extension}},
  booktitle = {SC24-W: Workshops of the International Conference for High Performance Computing, Networking, Storage and Analysis},
  pages = {1443--1454},
  year = {2024},
  doi = {10.1109/SCW63240.2024.00185}
}

@inproceedings{reddi2020mlperf,
  author = {Vijay Janapa Reddi and Christine Cheng and David Kanter and Peter Mattson
            and Guenther Schmuelling and Carole-Jean Wu and Brian Anderson
            and Maximilien Breughe and Mark Charlebois and William Chou
            and Ramesh Chukka and Cody Coleman and Sam Davis and Pan Deng
            and Greg Diamos and Jared Duke and Dave Fick and J. Scott Gardner
            and Itay Hubara and Sachin Idgunji and Thomas B. Jablin and Jeff Jiao
            and Tom St. John and Pankaj Kanwar and David Lee and Jeffery Liao
            and Anton Lokhmotov and Francisco Massa and Peng Meng
            and Paulius Micikevicius and Colin Osborne and Gennady Pekhimenko
            and Arun Tejusve Raghunath Rajan and Dilip Sequeira
            and Ashish Sirasao and Fei Sun and Hanlin Tang and Michael Thomson
            and Frank Wei and Ephrem Wu and Lingjie Xu and Koichi Yamada
            and Bing Yu and George Yuan and Aaron Zhong and Peizhao Zhang
            and Yuchen Zhou},
  title = {{MLPerf} Inference Benchmark},
  booktitle = {2020 ACM/IEEE 47th Annual International Symposium on Computer Architecture (ISCA)},
  pages = {446--459},
  year = {2020},
  doi = {10.1109/ISCA45697.2020.00045}
}

@article{chen2021codex,
  title={Evaluating Large Language Models Trained on Code},
  author={Chen, Mark and Tworek, Jerry and Jun, Heewoo and Yuan, Qiming and Pinto, Henrique Ponde de Oliveira and Kaplan, Jared and Edwards, Harri and Burda, Yuri and Joseph, Nicholas and Brockman, Greg and Ray, Alex and Puri, Raul and Krueger, Gretchen and Petrov, Michael and Khlaaf, Heidy and Sastry, Girish and Mishkin, Pamela and Chan, Brooke and Gray, Scott and Ryder, Nick and Pavlov, Mikhail and Power, Alethea and Kaiser, Lukasz and Bavarian, Mohammad and Winter, Clemens and Tillet, Philippe and Such, Felipe Petroski and Cummings, Dave and Plappert, Matthias and Chantzis, Fotios and Barnes, Elizabeth and Herbert-Voss, Ariel and Guss, William Hebgen and Nichol, Alex and Paino, Alex and Tezak, Nikolas and Tang, Jie and Babuschkin, Igor and Balaji, Suchir and Jain, Shantanu and Saunders, William and Hesse, Christopher and Carr, Andrew N. and Leike, Jan and Achiam, Josh and Misra, Vedant and Morikawa, Evan and Radford, Alec and Knight, Matthew and Brundage, Miles and Murati, Mira and Mayer, Katie and Welinder, Peter and McGrew, Bob and Amodei, Dario and McCandlish, Sam and Sutskever, Ilya and Zaremba, Wojciech},
  journal={arXiv preprint arXiv:2107.03374},
  year={2021}
}

@article{austin2021mbpp,
  title={Program Synthesis with Large Language Models},
  author={Austin, Jacob and Odena, Augustus and Nye, Maxwell and Bosma, Maarten and Michalewski, Henryk and Dohan, David and Jiang, Ellen and Cai, Carrie and Terry, Michael and Le, Quoc and Sutton, Charles},
  journal={arXiv preprint arXiv:2108.07732},
  year={2021}
}

@inproceedings{jain2025livecodebench,
  title={{LiveCodeBench}: Holistic and Contamination Free Evaluation of Large Language Models for Code},
  author={Jain, Naman and Han, King and Gu, Alex and Li, Wen-Ding and Yan, Fanjia and Zhang, Tianjun and Wang, Sida and Solar-Lezama, Armando and Sen, Koushik and Stoica, Ion},
  booktitle={International Conference on Learning Representations (ICLR)},
  year={2025}
}

@inproceedings{jimenez2024swebench,
  title={{SWE}-bench: Can Language Models Resolve Real-World {GitHub} Issues?},
  author={Jimenez, Carlos E. and Yang, John and Wettig, Alexander and Yao, Shunyu and Pei, Kexin and Press, Ofir and Narasimhan, Karthik},
  booktitle={International Conference on Learning Representations (ICLR)},
  year={2024}
}

@inproceedings{yang2024sweagent,
  title={{SWE}-agent: Agent-Computer Interfaces Enable Automated Software Engineering},
  author={Yang, John and Jimenez, Carlos E. and Wettig, Alexander and Lieret, Kilian and Yao, Shunyu and Narasimhan, Karthik and Press, Ofir},
  booktitle={Advances in Neural Information Processing Systems (NeurIPS)},
  year={2024}
}

@inproceedings{wang2025openhands,
  title={{OpenHands}: An Open Platform for {AI} Software Developers as Generalist Agents},
  author={Wang, Xingyao and Li, Boxuan and Song, Yufan and Xu, Frank F. and Tang, Xiangru and Zhuge, Mingchen and Pan, Jiayi and Song, Yueqi and Li, Bowen and Singh, Jaskirat and Tran, Hoang H. and Li, Fuqiang and Ma, Ren and Zheng, Mingzhang and Qian, Bill and Shao, Yanjun and Muennighoff, Niklas and Zhang, Yizhe and Hui, Binyuan and Lin, Junyang and Brennan, Robert and Peng, Hao and Ji, Heng and Neubig, Graham},
  booktitle={International Conference on Learning Representations (ICLR)},
  year={2025}
}

@misc{nanobot,
  title={nanobot: Ultra-Lightweight Open-Source Personal {AI} Agent Framework},
  author={{HKUDS}},
  year={2025},
  howpublished={\url{https://github.com/HKUDS/nanobot}},
  note={Accessed 2026-09-14}
}

@misc{anthropic2025claudecode,
  title={{Claude Code}},
  author={{Anthropic}},
  year={2025},
  howpublished={\url{https://github.com/anthropics/claude-code}},
  note={Accessed 2026-09-14}
}

@inproceedings{he2026sweperf,
  title={{SWE-Perf}: Can Language Models Optimize Code Performance on Real-World Repositories?},
  author={He, Xinyi and Liu, Qian and Du, Mingzhe and Yan, Lin and Fan, Zhijie and Huang, Yiming and Zheng, Yin and Yuan, Zejian and Ma, Zejun},
  booktitle={International Conference on Machine Learning (ICML)},
  year={2026}
}

@inproceedings{ma2026swefficiency,
  title={{SWE-fficiency}: Can Language Models Optimize Real-World Repositories on Real Workloads?},
  author={Ma, Jeffrey Jian and Hashemi, Milad and Yazdanbakhsh, Amir and Swersky, Kevin and Press, Ofir and Li, Enhui and Reddi, Vijay Janapa and Ranganathan, Parthasarathy},
  booktitle={International Conference on Machine Learning (ICML)},
  year={2026}
}

@article{garg2025perfbench,
  title={{PerfBench}: Can Agents Resolve Real-World Performance Bugs?},
  author={Garg, Spandan and Zilouchian Moghaddam, Roshanak and Sundaresan, Neel},
  journal={arXiv preprint arXiv:2509.24091},
  year={2025}
}

@article{cui2026perfoptbench,
  title={{PERFOPT-Bench}: Evaluating Coding Agents on Software Performance Optimization},
  author={Cui, Yingyun and Xie, Yi and Wang, Piaohong and Ma, Jiawei and Liu, Bo and Cao, Liangliang},
  journal={arXiv preprint arXiv:2607.07744},
  year={2026}
}

@inproceedings{ouyang2025kernelbench,
  title={{KernelBench}: Can {LLMs} Write Efficient {GPU} Kernels?},
  author={Ouyang, Anne and Guo, Simon and Arora, Simran and Zhang, Alex L. and Hu, William and R{\'e}, Christopher and Mirhoseini, Azalia},
  booktitle={International Conference on Machine Learning (ICML)},
  year={2025}
}

@article{wen2025multikernelbench,
  title={{MultiKernelBench}: A Multi-Platform Benchmark for Kernel Generation},
  author={Wen, Zhongzhen and Zhang, Yinghui and Li, Zhong and Liu, Zhongxin and Xie, Linna and Zhang, Tian},
  journal={arXiv preprint arXiv:2507.17773},
  year={2025}
}

@article{zang2026kernelgenbench,
  title={{KernelGenBench}: Can {LLMs} and Agents Write Efficient Kernels Across Operator Sources and Hardware Platforms?},
  author={Zang, Peiyu and Tao, Jian and Zhang, Jialing and Yuan, Yichen and Zhang, Wentao and Liu, Guang and Lin, Yonghua},
  journal={arXiv preprint arXiv:2607.27231},
  year={2026}
}

@article{zou2026mobilekernelbench,
  title={{MobileKernelBench}: Can {LLMs} Write Efficient Kernels for Mobile Devices?},
  author={Zou, Xingze and Wang, Jing and Zheng, Yuhua and Chen, Xueyi and Bai, Haolei and Kong, Lingcheng and Abu-Bakar, Syed A. R. and Wang, Zhaode and Lv, Chengfei and Hu, Haoji and Wang, Huan},
  journal={arXiv preprint arXiv:2603.11935},
  year={2026}
}

@article{li2026codegenbench,
  title={{CodegenBench}: Can {LLMs} Write Efficient Code Across Architectures?},
  author={Li, Jie and Wu, Wenzhao and Hu, Junqi and Zheng, Qinrui and Wu, Bowen and Zheng, Juepeng and Lu, Yutong and Fu, Haohuan},
  journal={arXiv preprint arXiv:2606.04023},
  year={2026}
}

@article{chen2025eok,
  title={Evolution of Kernels: Automated {RISC-V} Kernel Optimization with Large Language Models},
  author={Chen, Siyuan and Lu, Zhichao and Zhang, Qingfu},
  journal={arXiv preprint arXiv:2509.14265},
  year={2025}
}

@inproceedings{li2025tritonbench,
  title={{TritonBench}: Benchmarking Large Language Model Capabilities for Generating {Triton} Operators},
  author={Li, Jianling and Li, Shangzhan and Gao, Zhenye and Shi, Qi and Li, Yuxuan and Wang, Zefan and Huang, Jiacheng and Wang, Haojie and Wang, Jianrong and Han, Xu and Liu, Zhiyuan and Sun, Maosong},
  booktitle={Findings of the Association for Computational Linguistics: ACL 2025},
  year={2025}
}

@inproceedings{andrews2025gpukernelscientist,
  title={{GPU} Kernel Scientist: An {LLM}-Driven Framework for Iterative Kernel Optimization},
  author={Andrews, Martin and Witteveen, Sam},
  booktitle={ES-FoMo Workshop at ICML},
  year={2025}
}

@article{zhang2025cudaforge,
  title={{CudaForge}: An Agent Framework with Hardware Feedback for {CUDA} Kernel Optimization},
  author={Zhang, Zijian and Wang, Rong and Li, Shiyang and Luo, Yuebo and Hong, Mingyi and Ding, Caiwen},
  journal={arXiv preprint arXiv:2511.01884},
  year={2025}
}

@article{dai2026cudaagent,
  title={{CUDA} Agent: Large-Scale Agentic {RL} for High-Performance {CUDA} Kernel Generation},
  author={Dai, Weinan and Wu, Hanlin and Yu, Qiying and Gao, Huan-ang and Li, Jiahao and Jiang, Chengquan and Lou, Weiqiang and Song, Yufan and Yu, Hongli and Chen, Jiaze and Ma, Wei-Ying and Zhang, Ya-Qin and Liu, Jingjing and Wang, Mingxuan and Liu, Xin and Zhou, Hao},
  journal={arXiv preprint arXiv:2602.24286},
  year={2026}
}

@article{lin2026solexecbench,
  title={{SOL-ExecBench}: Speed-of-Light Benchmarking for Real-World {GPU} Kernels Against Hardware Limits},
  author={Lin, Edward and Modi, Sahil and Hari, Siva Kumar Sastry and Huang, Qijing and Ye, Zhifan and Qin, Nestor and Zhou, Fengzhe and Zhang, Yuan and Wang, Jingquan and Damani, Sana and Peri, Dheeraj and Xie, Ouye and Kane, Aditya and Maor, Moshe and Behar, Michael and Cao, Triston and Mehta, Rishabh and Singh, Vartika and Mailthody, Vikram Sharma and Chen, Terry and Ye, Zihao and Chen, Hanfeng and Chen, Tianqi and Grover, Vinod and Chen, Wei and Liu, Wei and Chung, Eric and Ceze, Luis and Bringmann, Roger and Zeller, Cyril and Lightstone, Michael and Kozyrakis, Christos and Shi, Humphrey},
  journal={arXiv preprint arXiv:2603.19173},
  year={2026}
}

@inproceedings{chen2018tvm,
  title={{TVM}: An Automated End-to-End Optimizing Compiler for Deep Learning},
  author={Chen, Tianqi and Moreau, Thierry and Jiang, Ziheng and Zheng, Lianmin and Yan, Eddie and Cowan, Meghan and Shen, Haichen and Wang, Leyuan and Hu, Yuwei and Ceze, Luis and Guestrin, Carlos and Krishnamurthy, Arvind},
  booktitle={13th USENIX Symposium on Operating Systems Design and Implementation (OSDI)},
  year={2018}
}

@inproceedings{zheng2020ansor,
  title={Ansor: Generating High-Performance Tensor Programs for Deep Learning},
  author={Zheng, Lianmin and Jia, Chengfan and Sun, Minmin and Wu, Zhao and Yu, Cody Hao and Haj-Ali, Ameer and Wang, Yida and Yang, Jun and Zhuo, Danyang and Sen, Koushik and Gonzalez, Joseph E. and Stoica, Ion},
  booktitle={14th USENIX Symposium on Operating Systems Design and Implementation (OSDI)},
  year={2020}
}

@inproceedings{cummins2022compilergym,
  title={{CompilerGym}: Robust, Performant Compiler Optimization Environments for {AI} Research},
  author={Cummins, Chris and Wasti, Bram and Guo, Jiadong and Cui, Brandon and Ansel, Jason and Gomez, Sahir and Jain, Somya and Liu, Jia and Teytaud, Olivier and Steiner, Benoit and Tian, Yuandong and Leather, Hugh},
  booktitle={IEEE/ACM International Symposium on Code Generation and Optimization (CGO)},
  pages={92--105},
  year={2022}
}

@article{trofin2021mlgo,
  title={{MLGO}: a Machine Learning Guided Compiler Optimizations Framework},
  author={Trofin, Mircea and Qian, Yundi and Brevdo, Eugene and Lin, Zinan and Choromanski, Krzysztof and Li, David},
  journal={arXiv preprint arXiv:2101.04808},
  year={2021}
}

@article{wei2025supercoder,
  title={{SuperCoder}: Assembly Program Superoptimization with Large Language Models},
  author={Wei, Anjiang and Suresh, Tarun and Tan, Huanmi and Xu, Yinglun and Singh, Gagandeep and Wang, Ke and Aiken, Alex},
  journal={arXiv preprint arXiv:2505.11480},
  year={2025}
}

@article{he2025simdbench,
  title={{SimdBench}: Benchmarking Large Language Models for {SIMD}-Intrinsic Code Generation},
  author={He, Yibo and Zhao, Shuoran and Huang, Jiaming and Fu, Yingjie and Yu, Hao and Huang, Cunjian and Xie, Tao},
  journal={arXiv preprint arXiv:2507.15224},
  year={2025}
}

@article{li2026autoveccoder,
  title={{AutoVecCoder}: Teaching {LLMs} to Generate Explicitly Vectorized Code},
  author={Li, Shangzhan and Yin, Xinyu and Jin, Xuanyu and He, Ye and Zhou, Yuxin and Li, Yuxuan and Han, Xu and Che, Wanxiang and Shi, Qi and Liu, Ting and Sun, Maosong},
  journal={arXiv preprint arXiv:2605.17978},
  year={2026}
}

@article{mankowitz2023alphadev,
  title={Faster Sorting Algorithms Discovered Using Deep Reinforcement Learning},
  author={Mankowitz, Daniel J. and Michi, Andrea and Zhernov, Anton and Gelmi, Marco and Selvi, Marco and Paduraru, Cosmin and Leurent, Edouard and Iqbal, Shariq and Lespiau, Jean-Baptiste and Ahern, Alex and K{\"o}ppe, Thomas and Millikin, Kevin and Gaffney, Stephen and Elster, Sophie and Broshear, Jackson and Gamble, Chris and Milan, Kieran and Tung, Robert and Hwang, Minjae and Cemgil, Taylan and Barekatain, Mohammadamin and Li, Yujia and Mandhane, Amol and Hubert, Thomas and Schrittwieser, Julian and Hassabis, Demis and Kohli, Pushmeet and Riedmiller, Martin and Vinyals, Oriol and Silver, David},
  journal={Nature},
  volume={618},
  number={7964},
  pages={257--263},
  year={2023}
}

@article{fawzi2022alphatensor,
  title={Discovering Faster Matrix Multiplication Algorithms with Reinforcement Learning},
  author={Fawzi, Alhussein and Balog, Matej and Huang, Aja and Hubert, Thomas and Romera-Paredes, Bernardino and Barekatain, Mohammadamin and Novikov, Alexander and Ruiz, Francisco J. R. and Schrittwieser, Julian and Swirszcz, Grzegorz and Silver, David and Hassabis, Demis and Kohli, Pushmeet},
  journal={Nature},
  volume={610},
  number={7930},
  pages={47--53},
  year={2022}
}

@article{novikov2025alphaevolve,
  title={{AlphaEvolve}: A Coding Agent for Scientific and Algorithmic Discovery},
  author={Novikov, Alexander and V{\~u}, Ng{\^a}n and Eisenberger, Marvin and Dupont, Emilien and Huang, Po-Sen and Wagner, Adam Zsolt and Shirobokov, Sergey and Kozlovskii, Borislav and Ruiz, Francisco J. R. and Mehrabian, Abbas and Kumar, M. Pawan and See, Abigail and Chaudhuri, Swarat and Holland, George and Davies, Alex and Nowozin, Sebastian and Kohli, Pushmeet and Balog, Matej},
  journal={arXiv preprint arXiv:2506.13131},
  year={2025}
}

@article{stephens2017sve,
  title={The {ARM} {Scalable Vector Extension}},
  author={Stephens, Nigel and Biles, Stuart and Boettcher, Matthias and Eapen, Jacob and Eyole, Mbou and Gabrielli, Giacomo and Horsnell, Matt and Magklis, Grigorios and Martinez, Alejandro and Premillieu, Nathanael and Reid, Alastair and Rico, Alejandro and Walker, Paul},
  journal={IEEE Micro},
  volume={37},
  number={2},
  pages={26--39},
  year={2017}
}

@misc{armsimdloops,
  title={{SIMD Loops}},
  author={{Arm}},
  year={2025},
  howpublished={\url{https://gitlab.arm.com/architecture/simd-loops}},
  note={BSD-3-Clause. Accessed 2026-09-14}
}

@misc{ncnn,
  title={ncnn: A High-Performance Neural Network Inference Framework Optimized for the Mobile Platform},
  author={{Tencent}},
  year={2017},
  howpublished={\url{https://github.com/Tencent/ncnn}},
  note={Accessed 2026-09-14}
}

@misc{gerganov2023llamacpp,
  title={llama.cpp: {LLM} Inference in {C/C++}},
  author={Gerganov, Georgi and {ggml-org contributors}},
  year={2023},
  howpublished={\url{https://github.com/ggml-org/llama.cpp}},
  note={Accessed 2026-09-16}
}

@misc{kleidiai,
  title={{KleidiAI}},
  author={{Arm}},
  year={2024},
  howpublished={\url{https://gitlab.arm.com/kleidi/kleidiai}},
  note={Apache-2.0. Accessed 2026-09-25}
}

@misc{sharegpt,
  title={{ShareGPT\_Vicuna\_unfiltered}},
  author={{ShareGPT}},
  year={2023},
  howpublished={\url{https://huggingface.co/datasets/anon8231489123/ShareGPT_Vicuna_unfiltered}},
  note={File ShareGPT\_V3\_unfiltered\_cleaned\_split.json. Apache-2.0. Accessed 2026-09-25}
}

@misc{openai2025codexcli,
  title={{OpenAI Codex CLI}},
  author={{OpenAI}},
  year={2025},
  howpublished={\url{https://github.com/openai/codex}},
  note={Accessed 2026-09-21}
}

@misc{anthropic2026opus48,
  title={Introducing {Claude Opus 4.8}},
  author={{Anthropic}},
  year={2026},
  howpublished={\url{https://www.anthropic.com/news/claude-opus-4-8}},
  note={Accessed 2026-09-21}
}

@misc{anthropic2026sonnet46,
  title={Introducing {Claude Sonnet 4.6}},
  author={{Anthropic}},
  year={2026},
  howpublished={\url{https://www.anthropic.com/news/claude-sonnet-4-6}},
  note={Accessed 2026-09-21}
}

@misc{anthropic2026fable51,
    title={Introducing {Claude Fable 5.1} and {Claude Mythos 5.1}},
    author={{Anthropic}},        
    year={2026},
    howpublished={\url{https://www.anthropic.com/claude-fable-and-mythos-5-1}},
    note={Accessed 2026-09-25}
  }

@misc{deepseek2026v4,
  title={{DeepSeek-V4} Model Card},
  author={{DeepSeek-AI}},
  year={2026},
  howpublished={\url{https://fe-static.deepseek.com/chat/transparency/deepseek-V4-model-card-EN.pdf}},
  note={Accessed 2026-09-21}
}

@misc{zhipu2026glm,
  title={{GLM-5.3}},
  author={{Zhipu AI}},
  year={2026},
  howpublished={\url{https://huggingface.co/zai-org/GLM-5.3}},
  note={Accessed 2026-09-21}
}

@inproceedings{mendis2019revec,
  title={Revec: Program Rejuvenation through Revectorization},
  author={Mendis, Charith and Jain, Ajay and Jain, Paras and Amarasinghe, Saman},
  booktitle={Proceedings of the 28th International Conference on Compiler Construction (CC)},
  year={2019}
}

@article{du2026adaexplore,
  title={{AdaExplore}: Failure-Driven Adaptation and Diversity-Preserving Search for Efficient Kernel Generation},
  author={Du, Weihua and Zhuo, Jingming and Dong, Yixin and He, Andre Wang and Sun, Weiwei and Zheng, Zeyu and Karunaratne, Manupa and Fox, Ivan and Dettmers, Tim and Chen, Tianqi and Yang, Yiming and Welleck, Sean},
  journal={arXiv preprint arXiv:2604.16625},
  year={2026}
}

@article{fu2026davincikernel,
  title={{daVinci-kernel}: Co-Evolving Skill Selection, Summarization, and Utilization via {RL} for {GPU} Kernel Optimization},
  author={Fu, Dayuan and Jiang, Mohan and Wang, Tongyu and Yang, Dian and Hu, Jiarui and Liu, Liming and Hou, Jinlong and Liu, Pengfei},
  journal={arXiv preprint arXiv:2606.16497},
  year={2026}
}

@article{tschand2026jaxbench,
  title={{JAXBench}: Benchmarking Autonomous {TPU} Kernel Optimization},
  author={Tschand, Arya and Hong, Charles and Walker, Julian and Cai, Nina and Wang, Shangkun and Subramanian, Suvinay and Dev, Sundar and Janapa Reddi, Vijay and Yazdanbakhsh, Amir and Sankaran, Sethu},
  journal={arXiv preprint arXiv:2607.20466},
  year={2026}
}
